\documentclass{article}
\usepackage{graphicx} % Required for inserting images
\usepackage{float}
\usepackage{authblk}

\usepackage{algorithm}
\usepackage{algorithmic}
\usepackage{multirow}
\usepackage{booktabs}
\usepackage{float}
\usepackage{makecell}
\usepackage{makecell}
\usepackage{CJKutf8}
\usepackage[numbers,sort&compress]{natbib}
\usepackage[colorlinks=true, urlcolor=blue, linkcolor=red]{hyperref}
\title{Review of Large Vision Models and Visual Prompt Engineering }
\newcommand*\samethanks[1][\value{footnote}]{\footnotemark[#1]}

\author[1]{Jiaqi Wang}
\author[2]{Zhengliang Liu}
\author[2]{Lin Zhao}
\author[2]{Zihao Wu}
\author[3]{Chong Ma}
\author[1]{Sigang Yu}
\author[2]{Haixing Dai}
\author[9]{Qiushi Yang}
\author[4]{Yiheng Liu}
\author[3]{Songyao Zhang}
\author[1]{Enze Shi}
\author[11]{Yi Pan}
\author[3]{Tuo Zhang}
\author[12]{Dajiang Zhu}
\author[13]{Xiang Li}
\author[10]{Xi Jiang}
\author[4]{Bao Ge}
\author[8]{Yixuan Yuan}
\author[5,6,7]{Dinggang Shen}

\author[2]{Tianming Liu \thanks{Joint corresponding author: tianming.liu@gmail.com and shu.zhang@nwpu.edu.cn}}
\author[1]{Shu Zhang \samethanks}

\affil[1]{School of Computer Science, Northwestern Polytechnical University, Xi'an 710072, China}

\affil[2]{School of Computing, The University of Georgia, Athens 30602, USA}

\affil[3]{School of Automation, Northwestern Polytechnical University, Xi'an 710072, China}

\affil[4]{School of Physics and Information Technology, Shaanxi Normal University, Xi’an 710119 China}

\affil[5]{School of Biomedical Engineering, ShanghaiTech University, Shanghai 201210, China}

\affil[6]{Shanghai United Imaging Intelligence Co., Ltd., Shanghai 200230, China}

\affil[7]{Shanghai Clinical Research and Trial Center, Shanghai, 201210, China}

\affil[8]{Department of Electronic Engineering, Chinese University of Hong Kong, Hong Kong 999077, China}

\affil[9]{Department of Electronic Engineering, City University of Hong Kong, Hong Kong 999077, China}

\affil[10]{School of Life Science and Technology, University of Electronic Science and Technology of China, Chengdu 611731, China}

\affil[11]{Glasgow College, University of Electronic Science and Technology of China, Chengdu 611731, China}

\affil[12]{Department of Computer Science and Engineering, The University of Texas at Arlington, Arlington 76019, USA}

\affil[13]{Department of Radiology, Massachusetts General Hospital and Harvard Medical School, Boston 02115, USA}

\begin{document}
\maketitle
\begin{CJK}{UTF8}{gbsn}

\begin{abstract}

Visual prompt engineering is a fundamental methodology in the field of visual and image Artificial General Intelligence (AGI). As the development of large vision models progresses, the importance of prompt engineering becomes increasingly evident. Designing suitable prompts for specific visual tasks has emerged as a meaningful research direction. This review aims to summarize the methods employed in the computer vision domain for large vision models and visual prompt engineering, exploring the latest advancements in visual prompt engineering. We present influential large models in the visual domain and a range of prompt engineering methods employed on these models. It is our hope that this review provides a comprehensive and systematic description of prompt engineering methods based on large visual models, offering valuable insights for future researchers in their exploration of this field.

\end{abstract}

\maketitle

\section{Introduction}

Since the introduction of the Transformer architecture by Vaswani et al.~\cite{vaswani2017attention}, deep learning models have experienced remarkable advancements in both parameter size and complexity. Over time, the scale of these models has grown exponentially. Early examples of language models include BERT~\cite{devlin2018bert}, T5~\cite{raffel2020exploring}, GPT-1~\cite{radford2018improving}, GPT-2~\cite{radford2019language} and various BERT variants~\cite{liao2023mask,liu2019roberta}. In addition, there exists a multitude of domain-specific BERT variants that are tailored to optimize performance in distinct fields of study or industry~\cite{rezayi2022clinicalradiobert,rezayi2022agribert,liu2023context}. More recently, large language models~\cite{zhang2023comprehensive,wang2023large} have become building blocks for general-purpose AI systems and are typically trained on extensive datasets through self-supervised learning~\cite{liu2023summary,zhang2023comprehensive,wang2023large}. This exponential growth in the scale and complexity of these models has significantly enhanced their capacity to comprehend natural language, allowing them to adapt to various downstream tasks~\cite{holmes2023evaluating,liu2023deid,ma2023impressiongpt,wu2023exploring,zhong2023chatabl,liao2023differentiate}. Notable examples include GPT-3~\cite{brown2020language}, ChatGPT~\cite{openaiIntroducingChatGPT}, GPT-4~\cite{openai2023gpt}, and others~\cite{zhao2023survey}, including domain-specific large language models~\cite{liu2023radiology}. This ability to generalize across multiple downstream tasks without explicit training, commonly known as zero-shot generalization, represents a groundbreaking advancement in the field~\cite{liu2023summary,zhao2023brain,dai2023ad,rezayi2023exploring,wang2023prompt}. 

Inspired by the success of pre-trained language models in natural language processing (NLP), researchers have ventured into exploring pre-trained visual models in the field of computer vision. These visual models are pre-trained on massive image datasets and possess the ability to understand the content of images and extract rich semantic information. Examples of pre-trained visual models include ViT~\cite{kim2021vilt}, Swin Transformer~\cite{liu2022swin}, VideoMAE V2~\cite{wang2023videomae} and others~\cite{chen2022mask,zhao2023metavit,xiao2023instruction,chen2022unified,ma2022rectify,yu2023core,lyu2022classification,yu2023gyri,yu2022disentangling,ding2022accurate}. By learning representations and features from a large amount of data, these models enable computers to more effectively comprehend and analyze images for diverse downstream applications~\cite{wang2023all,bi2023community,zhang2023beam,ding2023deep,luo2023towards,liu2022discovering,balagopal2021psa}. Moreover, multi-modal visual models, such as CLIP~\cite{radford2021learning} and ALIGN~\cite{cohen1997align}, employ contrastive learning to align textual and visual information. This alignment enables pre-trained models to effectively apply learned semantic information to the visual domain, facilitating efficient generalization in downstream tasks. However, despite their remarkable achievements, these models still face limitations in terms of their generalization capabilities.

The rapid advancements in artificial intelligence (AI) have given rise to a plethora of exciting technological breakthroughs, among which the development of AI systems based on foundational models has emerged as a prominent area of research~\cite{li2023artificial}. This conceptual framework has been coined and unified by AI experts, representing an emerging paradigm in the field~\cite{bommasani2021opportunities}. The significance of this concept extends to the notion of emergence, which has become increasingly evident with the rise of machine learning techniques. Emergence manifests itself in the execution of tasks such as automatic inference and the progressive emergence of advanced features and functionalities through deep learning, such as contextual learning. The concept of emergence emphasizes that system behavior is intricately induced rather than explicitly constructed, underscoring the dynamic nature of foundational models in the AI landscape.

Recently, the Segment Anything Model (SAM)~\cite{kirillov2023segment} has brought about a new trend in solving downstream tasks. Models with prompt engineering modules can solve a wide range of downstream tasks through prompts~\cite{kirillov2023segment,zhang2023segment,zhang2023comprehensive}. These models' remarkable zero-shot generalization capability highlights the significance of prompt engineering in downstream tasks~\cite{ramesh2021zero}. However, applying large visual models to specific tasks necessitates an effective approach to guide the model's learning and inference processes~\cite{zhang2023comprehensive,wang2023large,kirillov2023segment}. This is where visual prompt engineering comes into play. It is a methodology that involves designing and optimizing visual prompts to steer large models toward generating the desired outputs.

The emergence of foundation models has unleashed tremendous potential for the advancement of artificial intelligence systems, with particular significance in the domain of computer vision. Visual prompt engineering serves as an adaptive interface and a versatile toolkit that seamlessly integrates with large visual models. By synergistically fusing the capabilities of large visual models with the ingenuity of visual prompt engineering, we empower ourselves to harness the full potential of foundation models, resulting in unparalleled flexibility and efficiency in the realm of image analysis and task resolution. This pioneering integration paves the way for exploring vast frontiers in artificial intelligence applications, unveiling many prospects and ushering in unprecedented opportunities.

\subsection{Scope and Focus of the Review}

This review focuses primarily on prompt engineering methods in computer vision. A collection of relevant literature was gathered by crawling arXiv using the keyword "visual prompt". As shown in Figure \ref{fig: trend}, articles unrelated to computer vision were subsequently filtered out using ChatGPT, resulting in a total account of 500 papers. These papers mainly discuss prompt algorithms pertinent to computer vision, ranging from multi-modal visual-language models to visual and general artificial intelligence models. Prompts take on various forms, including text prompts in multi-modal settings, image prompts, and text-image prompts, each requiring distinct characteristics for different tasks. This paper comprehensively reviews prompt engineering in computer vision, providing insights into different aspects such as multi-modal prompt design, image prompt design, and text-image prompt design, taking into account the specific requirements of different tasks. The aim is to shed light on the advancements and current state of prompt engineering in computer vision, thereby facilitating further research in this field.

\begin{figure}
	\centerline{\includegraphics[width=\columnwidth]{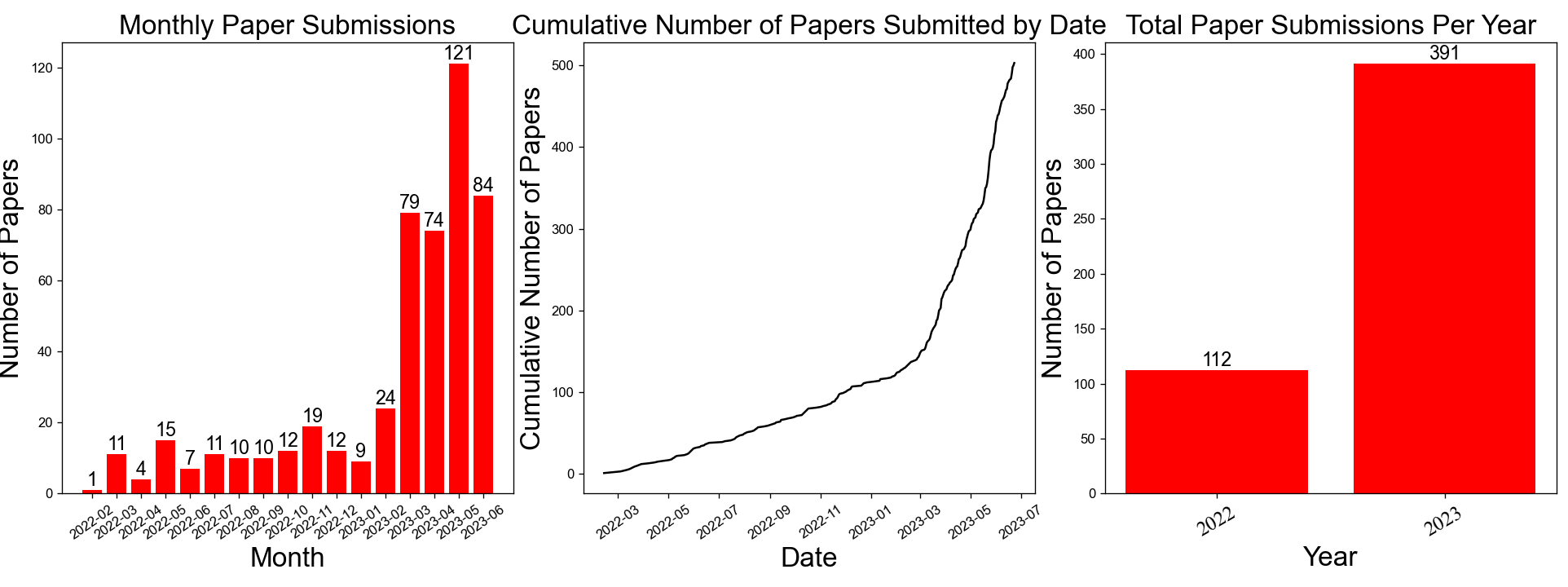}}
	\caption{The graphical representation is utilized to depict the number of research on viusal prompt engineering in the computer vision domain, published from 2022 to June 23, 2023, revealing the trend and growth of this field over time. The graph showcases three different plots: monthly submitted count, cumulative daily submitted count and annual submitted count.}
	\label{fig:  trend}
\end{figure}

\subsection{Outline of the Review}
This comprehensive review provides a scientific overview of the latest advancements in computer vision prompts and summarizes the existing design methods in the field. The review is structured as follows:

In the introduction, we traced the evolution of foundational AI models from the inception of the Transformer architecture to the development of large-scale vision models, highlighting how the growth and complexity of these models have spurred the innovative use of prompts (including visual prompts).

Section II presents an overview of key models that have contributed significantly to the advancement of visual prompts and AGI, including Transformer, CLIP, Visual Prompt Tuning (VPT), and SAM. These influential models serve as fundamental references for understanding the following discussions on prompt learning and application in AGI.

Section III delves into visual prompt learning, focusing on multi-modal prompts and visual prompt tuning. Different models and their variants specifically designed for multi-modal prompts are explored, highlighting different approaches and applications in this area. In addition, the section also discusses models and their variants that enable the effective tuning of visual prompts to enhance performance in specific tasks or domains, emphasizing the importance of selecting the appropriate prompt modality for different applications, as exemplified by the use of bounding box prompts in medical image segmentation and text prompts in natural image understanding.

Section IV focuses on the application of visual prompts in AGI models, highlighting the integration of prompts within AGI architectures and showcasing their contributions to strong generalization performance. The latest advancements in visual prompts for AGI models are presented, illustrating how proper prompt design can enable improved performance across diverse domains and tasks.

Section V explores future directions and implications of visual prompts research. We discuss potential developments in the field, taking into account advancements in AGI and related areas. Apart from advancements, the section also addresses the challenges and opportunities associated with them in visual prompts technology and provides insights into the broader implications and potential impact of visual prompts.

The conclusion section provides a summary of the main points discussed throughout the review. The critical role of visual prompts in AGI is emphasized, along with their potential for enhancing performance and generalization. The importance of prompt design and different modalities for different applications is reiterated, and future research prospects in this area are highlighted as well. The conclusion presents a concise recap of the significance of visual prompts and their implications for AGI, serving as a closing remark leaving readers with a clear understanding of the importance of visual prompts.

\section{Background Knowledge}

In this section, we will introduce some fundamental concepts, beginning with an elucidation of prompt engineering, followed by an overview of foundational models in the field of computer vision. Within this comprehensive review, our primary focus will be on the key techniques and important approaches pertaining to prompt engineering for large visual models.

\subsection{Prompts in Natural Language Processing}
In the field of NLP, to achieve parameter-efficient tuning for pre-trained models, prompt-based methods~\cite{prompt_nlp1,prompt_nlp2,prompt_nlp3,prompt_nlp4,prompt_nlp5,prompt_nlp6} are presented by montaging the inputs with additional context. Since prompt-based approaches can bridge the gap between pre-training and downstream tasks and unleash the potential of pre-trained models, they have witnessed remarkable superiority in various NLP tasks. 
According to the location of prompts within the text, prompts can be grouped into two shapes. The first is the cloze prompt, which exists in the middle of the text, and the other is the prefix prompt usually attaching at the end of the text.
Most previous works~\cite{prompt_nlp1,prompt_nlp2} design the desired prompts by manually defining or automatically learning.
The most simple manner to create the prompts is manually defining according to the common knowledge related to downstream tasks. Brown et. al.~\cite{brown2020language} manually define specific prompts towards multiple downstream NLP tasks including machine translation and question answering. Schick et. al.~\cite{prompt_nlp8} leverage pre-defined prompts to boost the few-shot text classification and generation tasks.
Although manually constructing prompts is simple and intuitive, they rely on complex different strategies for different tasks and need much professional experience, which are expensive and inefficient. Moreover, pre-designed prompts are usually not the optimal ones and they can't adaptively cope with many difficult tasks well. To yield more efficient prompt templates, many researches~\cite{prompt_nlp3,prompt_nlp5,prompt_nlp11,prompt_nlp10,prompt_nlp8} propose to automatically learn optimal prompts via sparse supervisions, which can be divided into two categories including the discrete prompts and continuous prompts. 
As the natural text context, discrete prompts are automatically searched pre-defined discrete space related to corresponding phrases. 
For example, Jiang et. al.~\cite{prompt_nlp8} present MINE as a mining-based method to automatically find prompts with both training inputs and outputs.  
Wallace et al.~\cite{prompt_nlp10} design a gradient-based search with input tokens to find short texts related to pre-trained models to generate the desired predictions in an iterative manner. Gao et al.~\cite{prompt_nlp11} regard the searching prompts as a sequence-to-sequence (seq2seq) generation task and leverage a seq2seq pre-trained model into the prompt searching process.
Instead of limit the prompts to natural language in the discrete space, other works~\cite{prompt_nlp3,prompt_nlp12,prompt_nlp13} aim to automatically construct desired prompts in the continuous text embedding space, which relaxes the searching scope and can be optimized via learnable parameters adaptively using downstream datasets.
For instance, Li et. al.~\cite{prompt_nlp12} prepend a sequence of continuous task-specific sequence to the inputs and keep the pre-trained models frozen. Lester et. al.~\cite{prompt_nlp13} prepend the inputs with special tokens to construct a prompt and explicitly tune the token embeddings.

% Tom B Brown, Benjamin Mann, Nick Ryder, Melanie Subbiah, Jared Kaplan, Prafulla Dhariwal, Arvind Neelakantan, Pranav Shyam, Girish Sastry, Amanda Askell, et al., 2020, "Language models are few-shot learners", arXiv preprint arXiv:2005.14165.
% Timo Schick and Hinrich Schutze. 2020. "Few-shot text generation with pattern-exploiting training".
% Zhengbao Jiang, Frank F. Xu, Jun Araki, and Graham Neubig. 2020c. How can we know what language models know? Transactions of the Association for Computational Linguistics, 8:423–438.
% Eric Wallace, Shi Feng, Nikhil Kandpal, Matt Gardner, and Sameer Singh. 2019a. Universal adversarial triggers for attacking and analyzing NLP. In Proceedings of the 2019 Conference on Empirical Methods in Natural Language Processing and the 9th International Joint Conference on Natural Language Processing, EMNLPIJCNLP 2019, Hong Kong, China, November 3-7, 2019, pages 2153–2162. Association for Computational Linguistics.
% Tianyu Gao, Adam Fisch, and Danqi Chen. 2021. Making pre-trained language models better few-shot learners. In Association for Computational Linguistics (ACL).
% Xiang Lisa Li and Percy Liang. 2021. Prefix-tuning: Optimizing continuous prompts for generation. arXiv preprint arXiv:2101.00190.
% Brian Lester, Rami Al-Rfou, and Noah Constant. 2021. The power of scale for parameter-efficient prompt tuning.

With prompts, the gap between pre-trained tasks and various downstream tasks can be narrowed, and performance can be efficiently boosted, potentially approaching the level of full parameter fine-tuning. This demonstrates that an appropriate parameter initialization can be considerably beneficial for downstream NLP tasks.

\subsection{Foundation Models}

\paragraph{Transformer} With the gradual development of large models, the Transformer architecture has emerged as a veritable foundational model, heralding a new era in the field. Transformer~\cite{vaswani2017attention} was first proposed for translation tasks in NLP, which combines Multi-head Self Attention (MSA) with Feed-forward Networks (FFN) to offer a global perceptual field and multi-channel feature extraction capabilities. The subsequent development of the Transformer-based BERT~\cite{kenton2019bert} proved to be seminal in NLP, exhibiting exceptional performance across multiple language-related tasks~\cite{cai2022coarse,liu2023deid,zhao2023brain}. Leveraging the great flexibility and scalability of the Transformer, researchers have started to train larger Transformer models, including GPT-1~\cite{radford2018improving}, GPT-2~\cite{radford2019language}, GPT-3~\cite{brown2020language}, GPT-4~\cite{openai2023gpt}, T5~\cite{raffel2020exploring}, PaLM~\cite{chowdhery2022palm}, LLaMA~\cite{touvron2023llama} and others. These models have further advanced the performance and generalization capabilities of the Transformer-based architecture, surpassing human-level performance in certain tasks~\cite{pan2023rewards}, and there is still potential for further development in terms of improving the effectiveness of training. Meanwhile, the Vision Transformer (ViT)~\cite{dosovitskiy2020image} has extended the application of Transformer architecture to the field of computer vision, bridging the gap between Transformer models in textual and image domains, and validating its feasibility as an unified architecture. Subsequent endeavors in computer vision began to improve and extend ViT model, such as DeiT~\cite{touvron2021training}, Swin Transformer~\cite{liu2021swin}, TNT~\cite{han2021transformer}, MAE~\cite{he2022masked}, MoCo-v3~\cite{chen2021empirical}, BeiT~\cite{bao2021beit}, etc. These works have successfully applied ViT to diverse vision-related tasks and achieved outstanding performance.

%\cite{liu2018generating,radford2019language,brown2020language,openai2023gpt4}
In the Transformer architecture~\cite{vaswani2017attention,dosovitskiy2020image}, the smallest unit of feature is a token. This inherent characteristic of the Transformer makes it well-suited for handling multi-modal data, as embedding layers can convert any modality into tokens. Consequently, numerous works in the multi-modal domain, such as ViTL~\cite{kim2021vilt}, DALL-E~\cite{ramesh2021zero}, CLIP~\cite{radford2021learning}, VLMO~\cite{bao2022vlmo}, ALBEF~\cite{li2021align} and others, have adopted the Transformer as the primary framework for multi-modal data interaction, including text-to-image and image-to-text retrieval, image captioning and image/text generation, etc. As the era of large models unfolds, researchers have proposed large multi-modal models such as CoCa~\cite{yu2022coca}, Flamingo~\cite{alayrac2022flamingo}, BEiT-v3~\cite{wang2023image}, PALI~\cite{chen2022pali}, GPT-4~\cite{openai2023gpt}, with the aim of further enhancing performance on a variety of downstream tasks. In summary, Transformer, as a fundamental model, continues to dominated today's research in the field.

\paragraph{CLIP} OpenAI has unveiled a groundbreaking vision-language model that leverages the association between images and text to perform weakly supervised pre-training, significantly boosting performance by expanding the available data. The developed work involves collecting a massive dataset of 400 million image-text pairs for training. CLIP~\cite{radford2021learning}, short for Contrastive Language-Image Pre-Training, utilizes fixed human-designed prompts that enable zero-shot prediction and demonstrates superior few-shot capabilities surpassing other state-of-the-art models. The success of CLIP highlights the power of combining visual and textual information and underscores the effectiveness of weakly supervised training using large data~\cite{khan2022transformers,croitoru2023diffusion,liang2023open}. CLIP's achievement signals a potential breakthrough in the understanding and application of multi-modal techniques, showcasing the ability to capture rich feature representations~\cite{saharia2022photorealistic}.such as Group VIT~\cite{xu2022groupvit}, ViLD~\cite{gu2021open}, Glip~\cite{li2022grounded}, Clipasso~\cite{vinker2022clipasso}, Clip4clip~\cite{luo2022clip4clip}, ActionCLIP~\cite{wang2021actionclip} and so on. It offers new insights into the advancement of prompt engineering in computer vision, providing a promising avenue for future developments~\cite{alayrac2022flamingo}.

\paragraph{VPT} When adapting large vision models to downstream tasks, modifying the input rather than altering the parameters of the pre-trained model itself is often preferred. This approach involves introducing a small number of task-specific learnable parameters into the input space, allowing for the learning of task-specific continuous vectors. This technique, known as prompt tuning, enables efficient fine-tuning of pre-trained models without modifying their underlying parameters. The VPT~\cite{jia2022visual} approach was the first to address and investigate the universality and feasibility of visual prompts. The proposed VPT method includes both deep and shallow versions, which attain impressive results by learning the prompt and class head of the input data while keeping the parameters of the pre-trained transformer model fixed. This work primarily aimed to demonstrate the effectiveness of visual prompting and provided a novel prompt design perspective. By showing that satisfactory results can be obtained by simply modifying the input, VPT showcased the potential of prompt tuning as an efficient strategy for downstream tasks and opened up new avenues for prompt engineering research~\cite{sohn2023visual}.

\paragraph{SAM} In 2023, Meta AI released a project aimed at creating a universal image segmentation model capable of addressing a wide range of downstream segmentation tasks on new data through prompt engineering. To achieve this, the SAM~\cite{kirillov2023segment} is created. SAM leverages prompt engineering to tackle general downstream segmentation tasks by utilizing the prompt segmentation task as a pre-training objective~\cite{deng2023segment}. To enhance the model's flexibility in adapting to prompts and to improve its robustness against interference, SAM is divided into three components: the image encoder, the prompt encoder, and the mask decoder. This division effectively distributes the computational cost, resulting in a sufficiently adaptable and versatile segmentation model~\cite{kirillov2023segment}. SAM’s strength lies in its ability to generalize efficiently across different segmentation tasks, thanks to the prompt engineering approach~\cite{zhang2023comprehensive,mazurowski2023segment,wu2023medical}. This methodology of pre-training on prompt segmentation and fine-tuning on specific downstream tasks helps SAM leverage the knowledge learned from the prompt segmentation task to improve performance on a wide range of segmentation problems, including medical image analysis~\cite{zhou2023can,shi2023generalist,he2023accuracy,zhang2023input}, video object tracking~\cite{cheng2023segment,yang2023track,yuan2023automated,cao2023ntire}, data annotation~\cite{julka2023knowledge,he2023scalable,he2023weakly}, 3D reconstruction~\cite{shen2023anything}, robotics~\cite{wang2023sam,diaz2023robot,beauchat2023analyzing}, image editing~\cite{yu2023inpaint,roy2023sam}, and more. Furthermore, SAM’s modular design allows for flexibility and adaptability to different prompt formats, making it a versatile solution for various segmentation challenges.

\section{Visual Prompts Learning}

\subsection{Multi-Modal Models and Prompts}

% \subsubsection{Vision-Language Models with Prompts}

The research field of multi-modal prompt learning has gained significant attention, with several notable works in the area. 

\paragraph{CLIP} One such work is CLIP~\cite{radford2021learning}, an innovative visual language model that incorporates the concept of manually crafted prompts. In CLIP, prompts take the form of "a photo of a [class]," with [class] denoting the specific data label. This design allows CLIP to comprehend both visual and textual information, establishing meaningful associations between these two modalities. However, fixed manual prompts have been found to be highly sensitive to results and have a significant impact on outcomes, an issue that has been noted in some studies. 

\paragraph{CoOP} To address this problem, CoOP~\cite{zhou2022learning} introduces the concept of automatic prompts. Automatic prompts represent the downstream task's prompt as a trainable continuous vector, enhancing prompt flexibility and adjustability. This approach allows prompts to be optimized based on specific task characteristics, rather than being limited to fixed manual settings. By training learnable prompt vectors, the CoOP model can automatically learn the appropriate prompt representation for different tasks. This flexibility enables the model to better adapt to diverse data and task requirements. However, the CoOP method exhibits lower generalization performance compared to CLIP on new data, which may be due to overfitting on downstream tasks. To address this issue, the authors introduce a lightweight Meta-Net that leverages the outputs of the image encoder and combines them with the trainable prompt. This results in a dynamic prompt that not only is a self-adaptive continuous vector learned regarding downstream tasks but also incorporates image features as conditions. The introduction of this dynamic prompt has significant implications for achieving better generalization performance.

\paragraph{DenseCLIP} DenseCLIP~\cite{rao2022denseclip} is a novel approach aimed at addressing the challenge of transferring large pre-trained models to dense tasks. While contrastive image-text pairing-based pre-training models demonstrate impressive performance on downstream tasks, the transferability to dense tasks remains a challenge for researchers that has yet to be explored. To address this gap, the researchers introduce an innovative method that incorporates CLIP and prompt patterns into dense tasks for the first time, along with a context-aware prompt that can adaptively adjust based on the specific task and input context. The utilization of context-aware prompts enables better capture of the semantic correlation between images and text in dense tasks, transforming the image-text matching problem into a pixel-text matching problem that improves model performance. DenseCLIP leverages large pre-trained models, such as CLIP, to learn the contrastive relationship between images and text, optimizing the model by maximizing the similarity between matched image-text pairs. This transformation and training strategy allows the model to better comprehend the fine-grained relationship between images and text in dense tasks. The introduction of DenseCLIP provides an innovative approach to transferring large pre-trained models to dense tasks. This method combines context-aware prompts and pixel-text matching problems, offering valuable insights and techniques for addressing the image-text correlation challenge in dense tasks.

\paragraph{MaPLe} The aforementioned work has demonstrated a series of significant advancements in the field of natural language processing, starting from manual text prompts to continuous text prompts and further extending to text-image prompts that incorporate image features. However, relying on prompts from a single modality results in suboptimal model performance. To address this issue, a new approach known as Multi-modal Prompt Learning (MaPLe) ~\cite{khattak2023maple} is proposed, where continuous prompts are used concurrently across multiple modalities. This approach emphasizes the interplay between text and images in prompt construction, enabling models to be enhanced to a certain extent. This dynamic method of prompt construction facilitates interactions and mutual influences between text and image prompts during model training. The introduction of this approach is significant for achieving more accurate and comprehensive multi-modal understanding. It leverages the interrelated information between text and images and incorporates more contextual and semantic information during the model training process, which further improves model performance.

\paragraph{Imagic} Imagic~\cite{kawar2023imagic} has proposed a groundbreaking framework for text-guided image editing, introducing complex text-based semantic editing for individual real-world images. For the first time, this framework enables the manipulation of object poses and compositions within an image while preserving their original features. By providing both an original image and a target text prompt, Imagic's framework allows for precise modifications that align with the semantic context of the image.

\paragraph{GALIP} Generative Adversarial CLIPs (GALIP)~\cite{tao2023galip}, a novel framework, has been proposed to enable text-to-image generation. Building upon the intricate scene understanding abilities and image comprehension of CLIP, GALIP introduces CLIP Visual Encoder (CLIPViT) and a learnable mate discriminator (Mate-D) for adversarial training, harnessing the generalization capabilities of CLIP. Ultimately, the framework employs text-conditioned prompts to adapt to downstream tasks, enhancing the synthesis capabilities for complex images.

\paragraph{PTP} To address the limitations of previous visual language model pre-training frameworks in terms of their lack of visual grounding and localization abilities, a new paradigm called Position-Guided Text Prompting (PTP)~\cite{wang2023position} has been proposed. PTP introduces a novel approach by encouraging the model to predict objects within given blocks or regress the blocks corresponding to specific objects. This reformulates visual grounding tasks as fill-in-the-blank problems using the provided PTP. Research findings have demonstrated that incorporating the PTP module into several state-of-the-art visual language model pre-training frameworks has led to significant improvements in representative cross-modal learning architectures and benchmark performance.

% \subsubsection{Vision-Audio Models with Prompts}

\subsection{Visual Prompts}

% \subsubsection{Image with Prompts}

The utilization of visual prompts in computer vision tasks can be traced back to interactive segmentation, a technique that requires user input, such as clicks~\cite{xu2016deep,wang2018interactive,jang2019interactive,lin2020interactive,chen2021conditional}, bounding boxes~\cite{lempitsky2009image,wu2014milcut,rajchl2016deepcut}, or scribbles \cite{batra2010icoseg,bai2014error,lin2016scribblesup}, to guide the algorithm in accurately identifying object boundaries. These visual prompts provide valuable guidance to the segmentation process, enabling more precise and reliable results. In the context of few-shot image segmentation, the annotated support image can also be considered as another form of visual prompt~\cite{shaban2017one,dong2018few,wang2019panet}. By leveraging the information in the support image and its corresponding segmentation mask, the algorithm can generalize and adapt its segmentation capabilities to similar target images. Recently, inspired by the success of prompt engineering in Natural Language Processing (NLP), the computer vision domain has witnessed a series of advancements in utilizing prompts. Researchers have started exploring the idea of formulating prompts as continuous vectors tailored to specific visual tasks. This involves designing prompts as guiding signals for computer vision models, helping them to generate or analyze visual content more effectively. By fine-tuning the models with task-specific prompts, notable improvements have been achieved in various visual tasks, including image classification, object detection, and image generation.

\paragraph{VPT} In addition to various prompting techniques in different modes, innovative developments in visual image prompting in the field of computer vision have emerged dramatically. VPT~\cite{jia2022visual} draws inspiration from large NLP models and employs prompt engineering to guide the fine-tuning process on a frozen pre-trained backbone model. It achieves this by introducing a small number of trainable parameters in the input space to serve as prompts. By optimizing these prompts, VPT enhances model performance regarding specific visual patterns and task requirements during the inference process, as depicted in the diagram. VPT is efficient, adaptable, and applicable to a wide range of visual tasks. By utilizing well-designed prompts, researchers can guide the model to perform better on specific visual tasks.

\paragraph{AdaptFormer} Similarly, AdaptFormer~\cite{chen2022adaptformer}, an existing ViT-based model, has been optimized to improve its efficiency on action recognition benchmarks by integrating lightweight modules into its architecture. The design philosophy behind AdaptFormer is to utilize trainable modules that are tailored to the task's specific constraints within the pre-trained ViT model. The experimental results show that AdaptFormer outperforms fully fine-tuned models in action recognition tasks.

\paragraph{Convpass} Convpass~\cite{jie2022convolutional} is a methodology aimed at rapidly tailoring pre-trained ViT by implementing convolutional bypasses. The primary goal of Convpass is to reduce the computational expenses associated with fine-tuning while increasing the adaptability of pre-trained models for particular computer vision tasks. Convolutional bypasses, integrated in Convpass, permit accelerated training and inference of models, without compromising performance. This technique aims to simplify the learning process and maximize the efficiency of utilizing pre-trained ViT for targeted computer vision applications.

\paragraph{ViPT} ViPT~\cite{zhu2023visual} proposes a prompt-tuning approach to address the challenge of limited large data in downstream multi-modal tracking tasks. This technique allows for the direct utilization of existing foundational knowledge for extracting RGB-modal features in RGB+auxiliary modality tracking tasks. The prompt-tuning method involves fine-tuning the parameters of the base model while retaining the knowledge from the pre-training phase. The prompt module allows for flexible adaptation to task-specific data. ViPT tackles the issue of insufficient data in multi-modal tracking tasks by employing the prompt-tuning method. This approach leverages prior knowledge embedded in the pre-trained model to facilitate the extraction of RGB-modal features. By fine-tuning the base model while preserving its existing knowledge, the prompt module enables efficient adaptation to the specific data requirements of the task at hand. ViPT's prompt-tuning technique is an effective solution to overcome data scarcity in multi-modal tracking tasks, optimizing the base model's parameters, retaining valuable knowledge from pre-training, and offering adaptability to task-specific data. This approach ensures logical and fluent integration of RGB-modal features, maximizing performance in RGB+auxiliary modality tracking tasks.
% \subsubsection{Video with Prompts}

\paragraph{DAM-VP}To address the challenge of handling complex distribution shifts from the original pre-training data distribution when using a single dataset-specific prompt,  Diversity-Aware Meta Visual Prompting
(DAM-VP)~\cite{huang2023diversity} introduces the concept of diversity-aware meta visual prompts. This approach employs a diversity-adaptive mechanism to cluster the downstream dataset into smaller, homogeneous subsets, each with its own individually optimized prompt. The aim is to tackle the difficulties arising from transferring knowledge between different data distributions. Research has revealed that leveraging prompt knowledge learned from previous datasets can expedite convergence and improve performance on new datasets. The integration of diversity-aware meta visual prompts in DAM-VP enables the model to adaptively exploit the diversity within the downstream dataset, facilitating more effective transfer learning and improved generalization.

\section{Visual Prompts in AGI}

With the impressive generalization capabilities demonstrated by universal models across various domains, significant progress has been made in large computer vision models as well. By training base models on diverse datasets, these models are capable of adapting to downstream tasks through prompt learning. This approach not only alleviates training demands and optimizes resource utilization, but also introduces new avenues for the development of computer vision. For example, the innovative "segment anything model" achieves powerful zero-shot transfer capabilities for downstream tasks by employing appropriate prompts, hence its versatile applications in various domains. This universal artificial intelligence model can learn general concepts and exhibit zero-shot transfer capabilities on unknown data on account of its transferability, showcasing the immense potential of general artificial intelligence. Nevertheless, prompt engineering is still pivotal to generalizing the model to new tasks and cannot be overwhelmed by other factors. Prompt engineering refers to the process of designing prompts that enable the model to adapt and generalize to different tasks. The question is crucial as a properly designed prompt can lead to effective representation learning, thereby enhancing its performance on various tasks. The prompt should be emphasized as the key to guiding models to understand and accommodate the context and requirements of either a seen or unseen task.

Many other models have emerged, including notable models such as OneFormer~\cite{jain2023oneformer}, SegGPT~\cite{wang2023seggpt}, SEEM~\cite{zou2023segment}, Uni-Perceiver v2~\cite{li2023uni},  demonstrating powerful capabilities in general artificial intelligence and providing new possibilities for addressing various tasks. These models employ zero-shot transfer methods, making prompt learning a critical manifestation of model generalization. 

% For instance, the use of prompts such as points or boxes, scribbles, masks, texts, audio, and referred regions of another image leads to prompting the segmentation of regions. These advancements have propelled the development of segmentation tasks and offered greater flexibility and adaptability for model applications across different domains and tasks.

In this section, we will delve into a detailed explanation of the prompt construction methods based on promptly, interactive models, encompassing key elements such as object detection, multi-modal fusion, and the combination of various models. These methods provide powerful tools for harnessing the generalization capabilities of large models, thereby offering feasible solutions for achieving downstream tasks.

\subsection{Object Detection}

In the task of object detection, the role of prompt-based methods in attaining the ability to generalize is of utmost importance to achieve general artificial intelligence. These methods are considered a fundamental basis for achieving such generalization. Nevertheless, despite SAM claiming its ability to segment any object, its practical application has been questioned. In particular, concerns have been raised regarding the efficacy of SAM for applications such as medical image segmentation, camouflage object detection, mirror and transparent object detection, and other similar scenarios. As a result, recent studies have concentrated on evaluating SAM's performance in various settings. These studies have demonstrated that point or box prompts are highly effective in various practical scenarios. SAM has achieved robust zero-shot performance in natural images, remote sensing, and medical imaging domains. However, its ability to generalize in complex application scenarios, particularly where semantic information is ambiguous or environments with low contrast, may not meet task requirements. Therefore, additional research is necessary to enhance SAM's performance in complex environments. In the realm of crater detection, SAM is leveraged for automated image segmentation. Subsequently, the shape of each segmented mask is assessed, and additional processing steps, such as filtering and boundary extraction, are carried out as well.

\paragraph{Object Counting} In the domain of object counting~\cite{ji2023segment,ma2023can}, researchers adopt SAM by employing bounding boxes as prompts to generate segmentation masks. Dense image features obtained from an image encoder are multiplied and averaged with a reference object's feature vector. Subsequently, a point grid consisting of 32 points per edge is employed as a prompt for segmentation. The resulting mask feature vector is obtained by multiplying and averaging it together with the dense features. Eventually, to determine the total count, the cosine similarity between the predicted mask and the reference example's feature vectors is calculated. When the cosine similarity exceeds a predefined threshold, the target object is regarded as recognized. By calculating all of the target objects, the total count is finally obtained.

\paragraph{Remote Sensing SAM}In the domain of remote sensing image segmentation~\cite{julka2023knowledge,zhang2023text2seg}, due to the top-down perspective of remote sensing images, objects within the scene can have arbitrary orientations. Consequently, a technique was proposed to use the Rotated Bounding Box (R-Box) minimum enclosing horizontal rectangle as guidance for SAM segmentation when designing prompts. For the mask prompt, it is defined as the corresponding area enclosed by the bounding box. Previous studies have also demonstrated the suitability and effectiveness of bounding boxes in designing prompts for efficient annotation purposes~\cite{wang2023scaling}.

\paragraph{SAM-Adapter}SAM-Adapter~\cite{chen2023sam} has been developed to infuse specialized domain knowledge into the original SAM model, thereby enhancing its capability to generalize across various downstream tasks. The integration has yielded promising outcomes. The adapter is engineered to acquire relevant knowledge and generate task-specific prompts at the preliminary stage. Notably, prompts can be outputted at each Transformer layer. By adopting this approach, prompts included in the segmented network significantly improves the performance of SAM in challenging tasks. The approach has achieved impressive results in disparate domains such as pseudocolor object detection, shadow detection, and medical image segmentation.

\subsection{Multi-modal Fusion}
In recent research, the introduction of textual information into visual models, such as images, paintings, frames, and videos, has significantly diversified and improved the fidelity of tasks. Inspired by the concept of generative models, the incorporation of textual prompts into CLIP has emerged as an effective approach.

\paragraph{Text2Seg} Text2Seg~\cite{zhang2023text2seg} introduces a vision-language model that relies on text prompts as the input. The model operates as follows: First, the text prompt serves as an input for Grounding DINO, which generates bounding boxes. These bounding boxes guide SAM in generating segmentation masks. Following this, the CLIP Surgery process then generates heatmaps using the text prompts, and the point prompts derived from these heatmaps are fed into SAM. Lastly, a similarity algorithm is applied to obtain the ultimate segmentation map.

\paragraph{SAMText} SAMText~\cite{he2023scalable} introduces a versatile methodology for generating segmentation masks aimed at scene text in images or video frames. The process initiates, once the input is provided, by extracting bounding box coordinates from a scene text detection model, using existing annotations. These extracted bounding box coordinates serve as prompts for SAM, which facilitates the subsequent generation of masks. If the bounding boxes exhibit orientation, SAMText computes their minimum bounding rectangles to obtain horizontal bounding boxes, which, in turn, serve as SAM's prompts for mask generation.

\paragraph{Caption Anything} Caption Anything~\cite{wang2023caption} has introduced a fundamental model-enhanced image captioning framework, which facilitates multi-modal control encompassing both visual and linguistic aspects. The framework seamlessly integrates SAM and ChatGPT, merging the visual and language modalities such that users can interactively model with the framework. During usage, users initially utilize various prompts, specifically points or bounding boxes, to flexibly control the input image, thereby enabling interactive user manipulation. The framework further refines the output instructions using large language models, ensuring effective alignment with the user's intended meaning and achieving significant consistency with the user's intent.

\paragraph{SAA+} Segment Any Anomaly + (SAA+)~\cite{cao2023segment} introduces a novel technique of zero-shot anomaly segmentation that utilizes hybrid prompt regularization to enhance the adaptability of existing foundational models. The proposed regularization prompt incorporates domain-specific expertise and contextual information from the target image, thereby reinforcing more robust prompt and facilitating more accurate identification of anomalous regions. In addition, many works have similarly concluded that incorporating domain expert knowledge as prior support might offer a potential solution to segmentation problems in complex scenes~\cite{wang2023scaling}.

\subsection{Combination of Various Models}

In complex scenarios, SAM's performance often lacks robustness, necessitating a new solution that combines interactive methods with efficient tools. The combination of these functionalities presents several potentials for a wide range of fields and exhibits outstanding performance in various tasks.

\paragraph{Inpaint Anything} Image inpainting, a pathological inverse problem that involves restoring missing or damaged parts of an image with visually plausible structures and textures, has been extensively studied in the field of computer vision. Inpaint Anything (IA)~\cite{yu2023inpaint} has proposed a conceptual pipeline based on the combination of various foundational models. By leveraging the strengths of these models, IA introduces three key functionalities in image inpainting: Remove Anything, Fill Anything, and Replace Anything. The pipeline follows a precise sequence, as shown in the diagram. Initially, a click prompt is employed to automatically segment designated regions, creating masks. Next, state-of-the-art inpainting models such as LaMa and Stable Diffusion (SD) are utilized to fill these masks, effectively completing the removal task. Following this step, a robust AI model like SD leverages a meticulously designed text prompt to generate the specific content required for filling or replacing the voids, facilitating the successful completion of the entire operation.

\paragraph{Edit Everything} Edit Everything~\cite{xie2023edit} introduces a generative system that combines SAM, CLIP, and SD to edit images guided by both image and text inputs. The original image is first segmented into multiple fragments using SAM. Then, the image-editing process is guided by text prompts such that it transforms the source image into the target image, aligning with the provided source and target prompts.

\paragraph{SAM-Track} SAM-Track~\cite{cheng2023segment} proposes a video segmentation framework that combines Grounding-DINO, DeAOT, and SAM to enable interactive and automated object tracking and segmentation across multiple modalities. The framework incorporates interactive prompts in the form of click-prompt, box-prompt, and text-prompt in the first frame of the video to guide SAM's segmentation process. Subsequently, text-prompts are subsequently utilized in the following frames for further result refinement. This versatile framework finds applications in a wide range of domains, including unmanned aerial vehicle technology, autonomous driving, medical imaging, augmented reality, and biological analysis.

\paragraph{Explain Any Concept} Explain Any Concept (EAC)~\cite{sun2023explain} proposes a novel approach to explain concepts based on three pipelines. While SAM excels in instance segmentation, its integration into Explainable AI poses the computational challenge of excessive complexity. EAC solves this by employing SAM for initial segmentation and introducing a surrogate model for efficient explanation. The first stage of the process employs SAM for instance segmentation, followed by the utilization of a surrogate model that approximates the target deep neural network. In the final stage, the trained network is applied to the results obtained in the first stage, facilitating the effective interpretation of the model's predictions.

\section{Future Directions and Implications}

With the continuous advancement of powerful large vision models, the significance of prompts within these models has become more prominent. Designing well-crafted prompts to effectively guide downstream tasks has emerged as a burgeoning avenue for tackling this issue. However, the performance of general artificial intelligence remains constrained by its reliance on domain-specific knowledge. To overcome this limitation, future endeavors should focus on expanding the breadth of knowledge by integrating diverse and comprehensive datasets, employing interdisciplinary methodologies, and fostering fruitful collaborations among experts from various domains. These endeavors will contribute to enhancing the capabilities of AI systems and addressing the challenges associated with leveraging prompts in a more holistic manner.

\subsection{Adaption of Large Vision Models}

Large vision models have emerged as a prominent trend in the field of AI, prompting the need to address the challenge of effectively adapting these models to downstream tasks. Several key techniques offer potential solutions in this regard.

Prompt fine-tuning, an essential tool in general artificial intelligence, serves as a crucial step in enabling models to better apply to downstream tasks~\cite{liu2022p,hu2022p3,jia2022visual,abdel2022dialect}. By designing suitable prompt examples and predefined inputs, models can be guided to better align with the target task, thus enhancing performance through fine-tuning for improved downstream task completion.

Reinforcement learning enables models to continuously learn and adjust their parameters by leveraging feedback signals obtained from experimentation and errors, thereby maximizing their performance~\cite{matsuo2022deep,ladosz2022exploration,wurman2022outracing,morse2022determinants}. When combined with prompt fine-tuning, reinforcement learning demonstrates outstanding effectiveness in optimizing adaptive model performance.

Adapter modules enable efficient adjustments for specific tasks within large models by introducing small, functional modules~\cite{thomas2022efficient,goel2022cross}. This approach selectively modifies only certain parts of the model's structure without requiring significant changes to the overall architecture. Incorporating adapter modules in prompt engineering, not only maintains the integrity of the larger model but also introduces task-specific functional structures, enabling more targeted prompt construction.

Knowledge distillation is a technique that transfers knowledge from a large model to a smaller one. The key lies in compactly representing the knowledge from the larger model and applying it to new tasks, preserving the essential performance and generalization capabilities while facilitating natural deployment in new environments~\cite{zhao2022decoupled,lin2022knowledge,chen2022knowledge}. While prompt fine-tuning is a natural choice for large models, the effectiveness of prompt engineering in smaller models remains an open question. Knowledge distillation can assist in applying prompts to small models by transferring the foundational knowledge and generalization abilities from large models, thereby achieving local deployment of smaller models.

Researchers have already devised a range of model adaptation methods, which have proven effective in various domains. These methods encompass techniques such as transfer learning, domain adaptation, and fine-tuning. While these approaches have their merits, they may not fully address the unique challenges posed by different domains. As the pursuit of effective model adaptation continues, the future holds promise for even greater achievements in applying models to specific tasks across various domains.

\subsection{Challenges and Considerations for Visual AGI}

Pioneered by advancements in the field of NLP, the development of a unified framework for large models is poised to become a critical undertaking in the future of visual models. However, in stark contrast to the progress made in NLP, the challenges faced by large models in the computer vision domain are more pronounced.

First and foremost, a key aspect of achieving AGI lies in interactive engagement with the environment and maximizing rewards~\cite{xie2023towards}. While NLP benefits from well-defined learning environments, enabling textual communication and task completion through multi-turn dialogues, the CV domain lacks a clear path and lacks interactive environments~\cite{moravec1988mind,brooks1990elephants}. Building a realistic environment for a CV proves exceedingly difficult due to the high costs and risks associated with human-agent interactions. On the other hand, constructing a virtual environment poses challenges when it comes to transferring trained agents to real-world scenarios~\cite{moravec1988mind,xie2023towards}.

Moreover, the image space exhibits stronger semantic sparsity, domain variations, and infinite granularity compared to the text space~\cite{xie2023towards}. Building upon the immense success of NLP, a foundation for achieving a unified paradigm in the CV domain has been established. Future research endeavors can take inspiration from the development of large NLP models, employing generative pre-training techniques combined with fine-tuning through instructions to achieve a unified approach in the CV field. Additionally, incorporating the capabilities of NLP, and applying multimodal techniques to large vision models can enable the fusion of language and images as an interactive mode for generative pre-training. This, in turn, opens up new avenues for human-machine interaction as a novel pre-training interactive mode.

\subsection{Applications Across Multiple Domains}

Visual prompts and large visual models have enabled significant progress in fields where visual understanding and analysis are critical. For example, the prompt-driven SAM has unlocked new opportunities in domains like medical imaging, agriculture, image editing, object detection, audio-visual localization, and beyond~\cite{ji2023segment}. In the medical domain, visual prompts such as segmentation masks, bounding boxes, and key points are used to help detect diseases, quantify the severity of lesions, and analyze medical scans~\cite{zhang2023segment,ma2023segment,mazurowski2023segment}. For the typical treatment sites in radiation oncology, Zhang et al. compared the Dice and Jaccard outcomes between clinical manual delineation and automatic segmentation using SAM with box prompt and proved SAM's robust generalization capabilities in automatic segmentation for radiotherapy~\cite{zhang2023segment}. In agriculture, visual prompts could be used to monitor crop growth, detect weeds or pests, and estimate crop yields~\cite{ji2023segment,lu2023agi}. Yang et al. assessed the zero-shot segmentation performance of SAM on representative chicken segmentation tasks and proved that SAM-based object tracking could provide valuable data on the behavior and movement patterns of broiler birds~\cite{yang2023sam}. 

By harnessing the advancements in LVM prompts, numerous fields stand to benefit from their integration. The ability to align language and visual data opens doors to improved medical diagnoses~\cite{wang2023prompt}, empowering healthcare providers with valuable insights from image-based information. Additionally, the flexibility of LVM prompts enables transformative applications in the realm of natural images, facilitating creative image manipulations and empowering users with powerful editing capabilities. Moreover, the utilization of prompts in video tracking introduces new possibilities for seamless human-machine interaction, allowing for enhanced object detection and precise audio-visual localization.

As research in the field progresses, the increasing utilization of LVM prompts is anticipated to revolutionize various industries and domains. The synergy between language and visual models, facilitated by prompts, paves the way for novel solutions, improved efficiency, and enhanced user experiences across multiple domains. In summary, visual prompts provide annotated data for visual understanding across domains. They give context and guidance, enhancing the ability of machines to interpret visual data.  Visual prompts have become an important tool for optimizing visual recognition systems with the increasing use of machine vision.

\section{Conclusion}

This review paper provides a comprehensive assessment of the remarkable advancements achieved in the domain of prompt engineering within the field of computer vision. It encompasses a detailed exploration of the design of visual prompts based on the ViT network architecture and the application of prompts leveraging AGI models. From a model-centric standpoint, the study delves into the positive effects of prompts on downstream tasks, highlighting their potential to inspire and enhance the emergent capabilities of large models.

Moreover, this article offers an in-depth discussion of the significance and performance of prompt engineering in various scenarios and domains, emphasizing its pivotal role in the field of computer vision. The article emphasizes the immense potential inherent in prompt engineering, holding promise for groundbreaking advancements in this discipline. 

Finally, the paper concludes by providing insights into future research avenues, highlighting the remarkable prospects of utilizing prompt engineering to completely revolutionize computer vision. This technique has the unparalleled potential of improving current models and enabling the creation of novel applications, thus enlightening further research into this area. Given the growing significance of prompts in computer vision in various fields, the outcomes of this study are highly relevant and timely.

\bibliography{mybib}

\begin{thebibliography}{100}

\bibitem{vaswani2017attention}
Ashish Vaswani, Noam Shazeer, Niki Parmar, Jakob Uszkoreit, Llion Jones,
  Aidan~N Gomez, {\L}ukasz Kaiser, and Illia Polosukhin.
\newblock Attention is all you need.
\newblock {\em Advances in neural information processing systems}, 30, 2017.

\bibitem{devlin2018bert}
Jacob Devlin, Ming-Wei Chang, Kenton Lee, and Kristina Toutanova.
\newblock Bert: Pre-training of deep bidirectional transformers for language
  understanding.
\newblock {\em arXiv preprint arXiv:1810.04805}, 2018.

\bibitem{raffel2020exploring}
Colin Raffel, Noam Shazeer, Adam Roberts, Katherine Lee, Sharan Narang, Michael
  Matena, Yanqi Zhou, Wei Li, and Peter~J Liu.
\newblock Exploring the limits of transfer learning with a unified text-to-text
  transformer.
\newblock {\em The Journal of Machine Learning Research}, 21(1):5485--5551,
  2020.

\bibitem{radford2018improving}
Alec Radford, Karthik Narasimhan, Tim Salimans, Ilya Sutskever, et~al.
\newblock Improving language understanding by generative pre-training.
\newblock 2018.

\bibitem{radford2019language}
Alec Radford, Jeffrey Wu, Rewon Child, David Luan, Dario Amodei, Ilya
  Sutskever, et~al.
\newblock Language models are unsupervised multitask learners.
\newblock {\em OpenAI blog}, 1(8):9, 2019.

\bibitem{liao2023mask}
Wenxiong Liao, Zhengliang Liu, Haixing Dai, Zihao Wu, Yiyang Zhang, Xiaoke
  Huang, Yuzhong Chen, Xi~Jiang, Dajiang Zhu, Tianming Liu, et~al.
\newblock Mask-guided bert for few shot text classification.
\newblock {\em arXiv preprint arXiv:2302.10447}, 2023.

\bibitem{liu2019roberta}
Yinhan Liu, Myle Ott, Naman Goyal, Jingfei Du, Mandar Joshi, Danqi Chen, Omer
  Levy, Mike Lewis, Luke Zettlemoyer, and Veselin Stoyanov.
\newblock Roberta: A robustly optimized bert pretraining approach.
\newblock {\em arXiv preprint arXiv:1907.11692}, 2019.

\bibitem{rezayi2022clinicalradiobert}
Saed Rezayi, Haixing Dai, Zhengliang Liu, Zihao Wu, Akarsh Hebbar, Andrew~H
  Burns, Lin Zhao, Dajiang Zhu, Quanzheng Li, Wei Liu, et~al.
\newblock Clinicalradiobert: Knowledge-infused few shot learning for clinical
  notes named entity recognition.
\newblock In {\em International Workshop on Machine Learning in Medical
  Imaging}, pages 269--278. Springer, 2022.

\bibitem{rezayi2022agribert}
Saed Rezayi, Zhengliang Liu, Zihao Wu, Chandra Dhakal, Bao Ge, Chen Zhen,
  Tianming Liu, and Sheng Li.
\newblock Agribert: knowledge-infused agricultural language models for matching
  food and nutrition.
\newblock In {\em Proceedings of the Thirty-First International Joint
  Conference on Artificial Intelligence}, volume~7, pages 5150--5156, 2022.

\bibitem{liu2023context}
Zhengliang Liu, Xinyu He, Lei Liu, Tianming Liu, and Xiaoming Zhai.
\newblock Context matters: A strategy to pre-train language model for science
  education.
\newblock {\em arXiv preprint arXiv:2301.12031}, 2023.

\bibitem{zhang2023comprehensive}
Chunhui Zhang, Li~Liu, Yawen Cui, Guanjie Huang, Weilin Lin, Yiqian Yang, and
  Yuehong Hu.
\newblock A comprehensive survey on segment anything model for vision and
  beyond.
\newblock {\em arXiv preprint arXiv:2305.08196}, 2023.

\bibitem{wang2023large}
Xiao Wang, Guangyao Chen, Guangwu Qian, Pengcheng Gao, Xiao-Yong Wei, Yaowei
  Wang, Yonghong Tian, and Wen Gao.
\newblock Large-scale multi-modal pre-trained models: A comprehensive survey.
\newblock {\em Machine Intelligence Research}, pages 1--36, 2023.

\bibitem{liu2023summary}
Yiheng Liu, Tianle Han, Siyuan Ma, Jiayue Zhang, Yuanyuan Yang, Jiaming Tian,
  Hao He, Antong Li, Mengshen He, Zhengliang Liu, et~al.
\newblock Summary of chatgpt/gpt-4 research and perspective towards the future
  of large language models.
\newblock {\em arXiv preprint arXiv:2304.01852}, 2023.

\bibitem{holmes2023evaluating}
Jason Holmes, Zhengliang Liu, Lian Zhang, Yuzhen Ding, Terence~T Sio, Lisa~A
  McGee, Jonathan~B Ashman, Xiang Li, Tianming Liu, Jiajian Shen, et~al.
\newblock Evaluating large language models on a highly-specialized topic,
  radiation oncology physics.
\newblock {\em arXiv preprint arXiv:2304.01938}, 2023.

\bibitem{liu2023deid}
Zhengliang Liu, Xiaowei Yu, Lu~Zhang, Zihao Wu, Chao Cao, Haixing Dai, Lin
  Zhao, Wei Liu, Dinggang Shen, Quanzheng Li, et~al.
\newblock Deid-gpt: Zero-shot medical text de-identification by gpt-4.
\newblock {\em arXiv preprint arXiv:2303.11032}, 2023.

\bibitem{ma2023impressiongpt}
Chong Ma, Zihao Wu, Jiaqi Wang, Shaochen Xu, Yaonai Wei, Zhengliang Liu, Lei
  Guo, Xiaoyan Cai, Shu Zhang, Tuo Zhang, et~al.
\newblock Impressiongpt: an iterative optimizing framework for radiology report
  summarization with chatgpt.
\newblock {\em arXiv preprint arXiv:2304.08448}, 2023.

\bibitem{wu2023exploring}
Zihao Wu, Lu~Zhang, Chao Cao, Xiaowei Yu, Haixing Dai, Chong Ma, Zhengliang
  Liu, Lin Zhao, Gang Li, Wei Liu, et~al.
\newblock Exploring the trade-offs: Unified large language models vs local
  fine-tuned models for highly-specific radiology nli task.
\newblock {\em arXiv preprint arXiv:2304.09138}, 2023.

\bibitem{zhong2023chatabl}
Tianyang Zhong, Yaonai Wei, Li~Yang, Zihao Wu, Zhengliang Liu, Xiaozheng Wei,
  Wenjun Li, Junjie Yao, Chong Ma, Xiang Li, et~al.
\newblock Chatabl: Abductive learning via natural language interaction with
  chatgpt.
\newblock {\em arXiv preprint arXiv:2304.11107}, 2023.

\bibitem{liao2023differentiate}
Wenxiong Liao, Zhengliang Liu, Haixing Dai, Shaochen Xu, Zihao Wu, Yiyang
  Zhang, Xiaoke Huang, Dajiang Zhu, Hongmin Cai, Tianming Liu, et~al.
\newblock Differentiate chatgpt-generated and human-written medical texts.
\newblock {\em arXiv preprint arXiv:2304.11567}, 2023.

\bibitem{brown2020language}
Tom Brown, Benjamin Mann, Nick Ryder, Melanie Subbiah, Jared~D Kaplan, Prafulla
  Dhariwal, Arvind Neelakantan, Pranav Shyam, Girish Sastry, Amanda Askell,
  et~al.
\newblock Language models are few-shot learners.
\newblock {\em Advances in neural information processing systems},
  33:1877--1901, 2020.

\bibitem{openaiIntroducingChatGPT}
{I}ntroducing {C}hat{G}{P}{T} --- openai.com.
\newblock \url{https://openai.com/blog/chatgpt}.
\newblock [Accessed 03-Jul-2023].

\bibitem{openai2023gpt}
OpenAI.
\newblock Gpt-4 technical report.
\newblock {\em arXiv}, pages 2303--08774, 2023.

\bibitem{zhao2023survey}
Wayne~Xin Zhao, Kun Zhou, Junyi Li, Tianyi Tang, Xiaolei Wang, Yupeng Hou,
  Yingqian Min, Beichen Zhang, Junjie Zhang, Zican Dong, et~al.
\newblock A survey of large language models.
\newblock {\em arXiv preprint arXiv:2303.18223}, 2023.

\bibitem{liu2023radiology}
Zhengliang Liu, Aoxiao Zhong, Yiwei Li, Longtao Yang, Chao Ju, Zihao Wu, Chong
  Ma, Peng Shu, Cheng Chen, Sekeun Kim, et~al.
\newblock Radiology-gpt: A large language model for radiology.
\newblock {\em arXiv preprint arXiv:2306.08666}, 2023.

\bibitem{zhao2023brain}
Lin Zhao, Lu~Zhang, Zihao Wu, Yuzhong Chen, Haixing Dai, Xiaowei Yu, Zhengliang
  Liu, Tuo Zhang, Xintao Hu, Xi~Jiang, et~al.
\newblock When brain-inspired ai meets agi.
\newblock {\em arXiv preprint arXiv:2303.15935}, 2023.

\bibitem{dai2023ad}
Haixing Dai, Yiwei Li, Zhengliang Liu, Lin Zhao, Zihao Wu, Suhang Song,
  Ye~Shen, Dajiang Zhu, Xiang Li, Sheng Li, et~al.
\newblock Ad-autogpt: An autonomous gpt for alzheimer's disease infodemiology.
\newblock {\em arXiv preprint arXiv:2306.10095}, 2023.

\bibitem{rezayi2023exploring}
Saed Rezayi, Zhengliang Liu, Zihao Wu, Chandra Dhakal, Bao Ge, Haixing Dai,
  Gengchen Mai, Ninghao Liu, Chen Zhen, Tianming Liu, et~al.
\newblock Exploring new frontiers in agricultural nlp: Investigating the
  potential of large language models for food applications.
\newblock {\em arXiv preprint arXiv:2306.11892}, 2023.

\bibitem{wang2023prompt}
Jiaqi Wang, Enze Shi, Sigang Yu, Zihao Wu, Chong Ma, Haixing Dai, Qiushi Yang,
  Yanqing Kang, Jinru Wu, Huawen Hu, et~al.
\newblock Prompt engineering for healthcare: Methodologies and applications.
\newblock {\em arXiv preprint arXiv:2304.14670}, 2023.

\bibitem{kim2021vilt}
Wonjae Kim, Bokyung Son, and Ildoo Kim.
\newblock Vilt: Vision-and-language transformer without convolution or region
  supervision.
\newblock In {\em International Conference on Machine Learning}, pages
  5583--5594. PMLR, 2021.

\bibitem{liu2022swin}
Ze~Liu, Han Hu, Yutong Lin, Zhuliang Yao, Zhenda Xie, Yixuan Wei, Jia Ning, Yue
  Cao, Zheng Zhang, Li~Dong, et~al.
\newblock Swin transformer v2: Scaling up capacity and resolution.
\newblock In {\em Proceedings of the IEEE/CVF conference on computer vision and
  pattern recognition}, pages 12009--12019, 2022.

\bibitem{wang2023videomae}
Limin Wang, Bingkun Huang, Zhiyu Zhao, Zhan Tong, Yinan He, Yi~Wang, Yali Wang,
  and Yu~Qiao.
\newblock Videomae v2: Scaling video masked autoencoders with dual masking.
\newblock In {\em Proceedings of the IEEE/CVF Conference on Computer Vision and
  Pattern Recognition}, pages 14549--14560, 2023.

\bibitem{chen2022mask}
Yuzhong Chen, Zhenxiang Xiao, Lin Zhao, Lu~Zhang, Haixing Dai, David~Weizhong
  Liu, Zihao Wu, Changhe Li, Tuo Zhang, Changying Li, et~al.
\newblock Mask-guided vision transformer (mg-vit) for few-shot learning.
\newblock {\em arXiv preprint arXiv:2205.09995}, 2022.

\bibitem{zhao2023metavit}
Lin Zhao, Hexin Dong, Ping Wu, Jiaying Lu, Le~Lu, Jingren Zhou, Tianming Liu,
  Li~Zhang, Ling Zhang, Yuxing Tang, et~al.
\newblock Metavit: Metabolism-aware vision transformer for differential
  diagnosis of parkinsonism with 18 f-fdg pet.
\newblock In {\em International Conference on Information Processing in Medical
  Imaging}, pages 132--144. Springer, 2023.

\bibitem{xiao2023instruction}
Zhenxiang Xiao, Yuzhong Chen, Lu~Zhang, Junjie Yao, Zihao Wu, Xiaowei Yu,
  Yi~Pan, Lin Zhao, Chong Ma, Xinyu Liu, et~al.
\newblock Instruction-vit: Multi-modal prompts for instruction learning in vit.
\newblock {\em arXiv preprint arXiv:2305.00201}, 2023.

\bibitem{chen2022unified}
Yuzhong Chen, Yu~Du, Zhenxiang Xiao, Lin Zhao, Lu~Zhang, David~Weizhong Liu,
  Dajiang Zhu, Tuo Zhang, Xintao Hu, Tianming Liu, et~al.
\newblock A unified and biologically-plausible relational graph representation
  of vision transformers.
\newblock {\em arXiv preprint arXiv:2206.11073}, 2022.

\bibitem{ma2022rectify}
Chong Ma, Lin Zhao, Yuzhong Chen, David~Weizhong Liu, Xi~Jiang, Tuo Zhang,
  Xintao Hu, Dinggang Shen, Dajiang Zhu, and Tianming Liu.
\newblock Rectify vit shortcut learning by visual saliency.
\newblock {\em arXiv preprint arXiv:2206.08567}, 2022.

\bibitem{yu2023core}
Xiaowei Yu, Lu~Zhang, Haixing Dai, Yanjun Lyu, Lin Zhao, Zihao Wu, David Liu,
  Tianming Liu, and Dajiang Zhu.
\newblock Core-periphery principle guided redesign of self-attention in
  transformers.
\newblock {\em arXiv preprint arXiv:2303.15569}, 2023.

\bibitem{lyu2022classification}
Yanjun Lyu, Xiaowei Yu, Dajiang Zhu, and Lu~Zhang.
\newblock Classification of alzheimer's disease via vision transformer:
  Classification of alzheimer's disease via vision transformer.
\newblock In {\em Proceedings of the 15th International Conference on PErvasive
  Technologies Related to Assistive Environments}, pages 463--468, 2022.

\bibitem{yu2023gyri}
Xiaowei Yu, Lu~Zhang, Haixing Dai, Lin Zhao, Yanjun Lyu, Zihao Wu, Tianming
  Liu, and Dajiang Zhu.
\newblock Gyri vs. sulci: Disentangling brain core-periphery functional
  networks via twin-transformer.
\newblock {\em arXiv preprint arXiv:2302.00146}, 2023.

\bibitem{yu2022disentangling}
Xiaowei Yu, Lu~Zhang, Lin Zhao, Yanjun Lyu, Tianming Liu, and Dajiang Zhu.
\newblock Disentangling spatial-temporal functional brain networks via
  twin-transformers.
\newblock {\em arXiv preprint arXiv:2204.09225}, 2022.

\bibitem{ding2022accurate}
Y~Ding, Z~Liu, H~Feng, J~Holmes, Y~Yang, N~Yu, T~Sio, S~Schild, B~Li, and
  W~Liu.
\newblock Accurate and efficient deep neural network based deformable image
  registration method in lung cancer.
\newblock In {\em MEDICAL PHYSICS}, volume~49, pages E148--E148. WILEY 111
  RIVER ST, HOBOKEN 07030-5774, NJ USA, 2022.

\bibitem{wang2023all}
Jinpeng Wang, Yixiao Ge, Rui Yan, Yuying Ge, Kevin~Qinghong Lin, Satoshi
  Tsutsui, Xudong Lin, Guanyu Cai, Jianping Wu, Ying Shan, et~al.
\newblock All in one: Exploring unified video-language pre-training.
\newblock In {\em Proceedings of the IEEE/CVF Conference on Computer Vision and
  Pattern Recognition}, pages 6598--6608, 2023.

\bibitem{bi2023community}
Xia-An Bi, Ke~Chen, Siyu Jiang, Sheng Luo, Wenyan Zhou, Zhaoxu Xing, Luyun Xu,
  Zhengliang Liu, and Tianming Liu.
\newblock Community graph convolution neural network for alzheimer’s disease
  classification and pathogenetic factors identification.
\newblock {\em IEEE Transactions on Neural Networks and Learning Systems},
  2023.

\bibitem{zhang2023beam}
Lian Zhang, Jason~M Holmes, Zhengliang Liu, Sujay~A Vora, Terence~T Sio,
  Carlos~E Vargas, Nathan~Y Yu, Sameer~R Keole, Steven~E Schild, Martin Bues,
  et~al.
\newblock Beam mask and sliding window-facilitated deep learning-based accurate
  and efficient dose prediction for pencil beam scanning proton therapy.
\newblock {\em arXiv preprint arXiv:2305.18572}, 2023.

\bibitem{ding2023deep}
Yuzhen Ding, Hongying Feng, Yunze Yang, Jason Holmes, Zhengliang Liu, David
  Liu, William~W Wong, Nathan~Y Yu, Terence~T Sio, Steven~E Schild, et~al.
\newblock Deep-learning based fast and accurate 3d ct deformable image
  registration in lung cancer.
\newblock {\em Medical Physics}, 2023.

\bibitem{luo2023towards}
Dezhao Luo, Jiabo Huang, Shaogang Gong, Hailin Jin, and Yang Liu.
\newblock Towards generalisable video moment retrieval: Visual-dynamic
  injection to image-text pre-training.
\newblock In {\em Proceedings of the IEEE/CVF Conference on Computer Vision and
  Pattern Recognition}, pages 23045--23055, 2023.

\bibitem{liu2022discovering}
Yiheng Liu, Enjie Ge, Mengshen He, Zhengliang Liu, Shijie Zhao, Xintao Hu,
  Dajiang Zhu, Tianming Liu, and Bao Ge.
\newblock Discovering dynamic functional brain networks via spatial and
  channel-wise attention.
\newblock {\em arXiv preprint arXiv:2205.09576}, 2022.

\bibitem{balagopal2021psa}
Anjali Balagopal, Howard Morgan, Michael Dohopolski, Ramsey Timmerman, Jie
  Shan, Daniel~F Heitjan, Wei Liu, Dan Nguyen, Raquibul Hannan, Aurelie Garant,
  et~al.
\newblock Psa-net: Deep learning--based physician style--aware segmentation
  network for postoperative prostate cancer clinical target volumes.
\newblock {\em Artificial Intelligence in Medicine}, 121:102195, 2021.

\bibitem{radford2021learning}
Alec Radford, Jong~Wook Kim, Chris Hallacy, Aditya Ramesh, Gabriel Goh,
  Sandhini Agarwal, Girish Sastry, Amanda Askell, Pamela Mishkin, Jack Clark,
  et~al.
\newblock Learning transferable visual models from natural language
  supervision.
\newblock In {\em International conference on machine learning}, pages
  8748--8763. PMLR, 2021.

\bibitem{cohen1997align}
Gerson~H Cohen.
\newblock Align: a program to superimpose protein coordinates, accounting for
  insertions and deletions.
\newblock {\em Journal of applied crystallography}, 30(6):1160--1161, 1997.

\bibitem{li2023artificial}
Xiang Li, Lu~Zhang, Zihao Wu, Zhengliang Liu, Lin Zhao, Yixuan Yuan, Jun Liu,
  Gang Li, Dajiang Zhu, Pingkuan Yan, et~al.
\newblock Artificial general intelligence for medical imaging.
\newblock {\em arXiv preprint arXiv:2306.05480}, 2023.

\bibitem{bommasani2021opportunities}
Rishi Bommasani, Drew~A Hudson, Ehsan Adeli, Russ Altman, Simran Arora, Sydney
  von Arx, Michael~S Bernstein, Jeannette Bohg, Antoine Bosselut, Emma
  Brunskill, et~al.
\newblock On the opportunities and risks of foundation models.
\newblock {\em arXiv preprint arXiv:2108.07258}, 2021.

\bibitem{kirillov2023segment}
Alexander Kirillov, Eric Mintun, Nikhila Ravi, Hanzi Mao, Chloe Rolland, Laura
  Gustafson, Tete Xiao, Spencer Whitehead, Alexander~C Berg, Wan-Yen Lo, et~al.
\newblock Segment anything.
\newblock {\em arXiv preprint arXiv:2304.02643}, 2023.

\bibitem{zhang2023segment}
Yichi Zhang and Rushi Jiao.
\newblock How segment anything model (sam) boost medical image segmentation?
\newblock {\em arXiv preprint arXiv:2305.03678}, 2023.

\bibitem{ramesh2021zero}
Aditya Ramesh, Mikhail Pavlov, Gabriel Goh, Scott Gray, Chelsea Voss, Alec
  Radford, Mark Chen, and Ilya Sutskever.
\newblock Zero-shot text-to-image generation.
\newblock In {\em International Conference on Machine Learning}, pages
  8821--8831. PMLR, 2021.

\bibitem{prompt_nlp1}
Edoardo~Maria Ponti, Goran Glava{\v{s}}, Olga Majewska, Qianchu Liu, Ivan
  Vuli{\'c}, and Anna Korhonen.
\newblock Xcopa: A multilingual dataset for causal commonsense reasoning.
\newblock {\em arXiv preprint arXiv:2005.00333}, 2020.

\bibitem{prompt_nlp2}
Hal Daum{\'e}~III and Eric Brill.
\newblock Web search intent induction via automatic query reformulation.
\newblock In {\em Proceedings of HLT-NAACL 2004: Short Papers}, pages 49--52,
  2004.

\bibitem{prompt_nlp3}
Bill~Yuchen Lin, Wangchunshu Zhou, Ming Shen, Pei Zhou, Chandra Bhagavatula,
  Yejin Choi, and Xiang Ren.
\newblock Commongen: A constrained text generation challenge for generative
  commonsense reasoning.
\newblock {\em arXiv preprint arXiv:1911.03705}, 2019.

\bibitem{prompt_nlp4}
Hector Levesque, Ernest Davis, and Leora Morgenstern.
\newblock The winograd schema challenge.
\newblock In {\em Thirteenth international conference on the principles of
  knowledge representation and reasoning}, 2012.

\bibitem{prompt_nlp5}
Christian Buck, Jannis Bulian, Massimiliano Ciaramita, Wojciech Gajewski,
  Andrea Gesmundo, Neil Houlsby, and Wei Wang.
\newblock Ask the right questions: Active question reformulation with
  reinforcement learning.
\newblock {\em arXiv preprint arXiv:1705.07830}, 2017.

\bibitem{prompt_nlp6}
Jake Snell, Kevin Swersky, and Richard Zemel.
\newblock Prototypical networks for few-shot learning.
\newblock In {\em Advances in neural information processing systems},
  volume~30, 2017.

\bibitem{prompt_nlp8}
Timo Schick and Hinrich Sch{\"u}tze.
\newblock Few-shot text generation with pattern-exploiting training.
\newblock {\em arXiv preprint arXiv:2012.11926}, 2020.

\bibitem{prompt_nlp11}
Tianyu Gao, Adam Fisch, and Danqi Chen.
\newblock Making pre-trained language models better few-shot learners.
\newblock {\em arXiv preprint arXiv:2012.15723}, 2020.

\bibitem{prompt_nlp10}
Eric Wallace, Shi Feng, Nikhil Kandpal, Matt Gardner, and Sameer Singh.
\newblock Universal adversarial triggers for attacking and analyzing nlp.
\newblock {\em arXiv preprint arXiv:1908.07125}, 2019.

\bibitem{prompt_nlp12}
Xiang~Lisa Li and Percy Liang.
\newblock Prefix-tuning: Optimizing continuous prompts for generation.
\newblock {\em arXiv preprint arXiv:2101.00190}, 2021.

\bibitem{prompt_nlp13}
Brian Lester, Joshua Yurtsever, Siamak Shakeri, and Noah Constant.
\newblock Reducing retraining by recycling parameter-efficient prompts.
\newblock {\em arXiv preprint arXiv:2208.05577}, 2022.

\bibitem{kenton2019bert}
Jacob Devlin Ming-Wei~Chang Kenton and Lee~Kristina Toutanova.
\newblock Bert: Pre-training of deep bidirectional transformers for language
  understanding.
\newblock In {\em Proceedings of NAACL-HLT}, pages 4171--4186, 2019.

\bibitem{cai2022coarse}
Homgmin Cai, Wenxiong Liao, Zhengliang Liu, Xiaoke Huang, Yiyang Zhang, Siqi
  Ding, Sheng Li, Quanzheng Li, Tianming Liu, and Xiang Li.
\newblock Coarse-to-fine knowledge graph domain adaptation based on
  distantly-supervised iterative training.
\newblock {\em arXiv preprint arXiv:2211.02849}, 2022.

\bibitem{chowdhery2022palm}
Aakanksha Chowdhery, Sharan Narang, Jacob Devlin, Maarten Bosma, Gaurav Mishra,
  Adam Roberts, Paul Barham, Hyung~Won Chung, Charles Sutton, Sebastian
  Gehrmann, et~al.
\newblock Palm: Scaling language modeling with pathways.
\newblock {\em arXiv preprint arXiv:2204.02311}, 2022.

\bibitem{touvron2023llama}
Hugo Touvron, Thibaut Lavril, Gautier Izacard, Xavier Martinet, Marie-Anne
  Lachaux, Timoth{\'e}e Lacroix, Baptiste Rozi{\`e}re, Naman Goyal, Eric
  Hambro, Faisal Azhar, et~al.
\newblock Llama: Open and efficient foundation language models.
\newblock {\em arXiv preprint arXiv:2302.13971}, 2023.

\bibitem{pan2023rewards}
Alexander Pan, Chan~Jun Shern, Andy Zou, Nathaniel Li, Steven Basart, Thomas
  Woodside, Jonathan Ng, Hanlin Zhang, Scott Emmons, and Dan Hendrycks.
\newblock Do the rewards justify the means? measuring trade-offs between
  rewards and ethical behavior in the machiavelli benchmark.
\newblock {\em arXiv preprint arXiv:2304.03279}, 2023.

\bibitem{dosovitskiy2020image}
Alexey Dosovitskiy, Lucas Beyer, Alexander Kolesnikov, Dirk Weissenborn,
  Xiaohua Zhai, Thomas Unterthiner, Mostafa Dehghani, Matthias Minderer, Georg
  Heigold, Sylvain Gelly, et~al.
\newblock An image is worth 16x16 words: Transformers for image recognition at
  scale.
\newblock In {\em International Conference on Learning Representations}, 2020.

\bibitem{touvron2021training}
Hugo Touvron, Matthieu Cord, Matthijs Douze, Francisco Massa, Alexandre
  Sablayrolles, and Herv{\'e} J{\'e}gou.
\newblock Training data-efficient image transformers \& distillation through
  attention.
\newblock In {\em International Conference on Machine Learning}, pages
  10347--10357. PMLR, 2021.

\bibitem{liu2021swin}
Ze~Liu, Yutong Lin, Yue Cao, Han Hu, Yixuan Wei, Zheng Zhang, Stephen Lin, and
  Baining Guo.
\newblock Swin transformer: Hierarchical vision transformer using shifted
  windows.
\newblock In {\em Proceedings of the IEEE/CVF International Conference on
  Computer Vision}, pages 10012--10022, 2021.

\bibitem{han2021transformer}
Kai Han, An~Xiao, Enhua Wu, Jianyuan Guo, Chunjing Xu, and Yunhe Wang.
\newblock Transformer in transformer.
\newblock {\em Advances in Neural Information Processing Systems},
  34:15908--15919, 2021.

\bibitem{he2022masked}
Kaiming He, Xinlei Chen, Saining Xie, Yanghao Li, Piotr Doll{\'a}r, and Ross
  Girshick.
\newblock Masked autoencoders are scalable vision learners.
\newblock In {\em Proceedings of the IEEE/CVF Conference on Computer Vision and
  Pattern Recognition}, pages 16000--16009, 2022.

\bibitem{chen2021empirical}
Xinlei Chen, Saining Xie, and Kaiming He.
\newblock An empirical study of training self-supervised vision transformers.
\newblock In {\em Proceedings of the IEEE/CVF International Conference on
  Computer Vision}, pages 9640--9649, 2021.

\bibitem{bao2021beit}
Hangbo Bao, Li~Dong, Songhao Piao, and Furu Wei.
\newblock Beit: Bert pre-training of image transformers.
\newblock {\em arXiv preprint arXiv:2106.08254}, 2021.

\bibitem{bao2022vlmo}
Hangbo Bao, Wenhui Wang, Li~Dong, Qiang Liu, Owais~Khan Mohammed, Kriti
  Aggarwal, Subhojit Som, Songhao Piao, and Furu Wei.
\newblock Vlmo: Unified vision-language pre-training with
  mixture-of-modality-experts.
\newblock {\em Advances in Neural Information Processing Systems},
  35:32897--32912, 2022.

\bibitem{li2021align}
Junnan Li, Ramprasaath Selvaraju, Akhilesh Gotmare, Shafiq Joty, Caiming Xiong,
  and Steven Chu~Hong Hoi.
\newblock Align before fuse: Vision and language representation learning with
  momentum distillation.
\newblock {\em Advances in neural information processing systems},
  34:9694--9705, 2021.

\bibitem{yu2022coca}
Jiahui Yu, Zirui Wang, Vijay Vasudevan, Legg Yeung, Mojtaba Seyedhosseini, and
  Yonghui Wu.
\newblock Coca: Contrastive captioners are image-text foundation models.
\newblock {\em arXiv preprint arXiv:2205.01917}, 2022.

\bibitem{alayrac2022flamingo}
Jean-Baptiste Alayrac, Jeff Donahue, Pauline Luc, Antoine Miech, Iain Barr,
  Yana Hasson, Karel Lenc, Arthur Mensch, Katherine Millican, Malcolm Reynolds,
  et~al.
\newblock Flamingo: a visual language model for few-shot learning.
\newblock {\em Advances in Neural Information Processing Systems},
  35:23716--23736, 2022.

\bibitem{wang2023image}
Wenhui Wang, Hangbo Bao, Li~Dong, Johan Bjorck, Zhiliang Peng, Qiang Liu, Kriti
  Aggarwal, Owais~Khan Mohammed, Saksham Singhal, Subhojit Som, et~al.
\newblock Image as a foreign language: Beit pretraining for vision and
  vision-language tasks.
\newblock In {\em Proceedings of the IEEE/CVF Conference on Computer Vision and
  Pattern Recognition}, pages 19175--19186, 2023.

\bibitem{chen2022pali}
Xi~Chen, Xiao Wang, Soravit Changpinyo, AJ~Piergiovanni, Piotr Padlewski,
  Daniel Salz, Sebastian Goodman, Adam Grycner, Basil Mustafa, Lucas Beyer,
  et~al.
\newblock Pali: A jointly-scaled multilingual language-image model.
\newblock {\em arXiv preprint arXiv:2209.06794}, 2022.

\bibitem{khan2022transformers}
Salman Khan, Muzammal Naseer, Munawar Hayat, Syed~Waqas Zamir, Fahad~Shahbaz
  Khan, and Mubarak Shah.
\newblock Transformers in vision: A survey.
\newblock {\em ACM computing surveys (CSUR)}, 54(10s):1--41, 2022.

\bibitem{croitoru2023diffusion}
Florinel-Alin Croitoru, Vlad Hondru, Radu~Tudor Ionescu, and Mubarak Shah.
\newblock Diffusion models in vision: A survey.
\newblock {\em IEEE Transactions on Pattern Analysis and Machine Intelligence},
  2023.

\bibitem{liang2023open}
Feng Liang, Bichen Wu, Xiaoliang Dai, Kunpeng Li, Yinan Zhao, Hang Zhang,
  Peizhao Zhang, Peter Vajda, and Diana Marculescu.
\newblock Open-vocabulary semantic segmentation with mask-adapted clip.
\newblock In {\em Proceedings of the IEEE/CVF Conference on Computer Vision and
  Pattern Recognition}, pages 7061--7070, 2023.

\bibitem{saharia2022photorealistic}
Chitwan Saharia, William Chan, Saurabh Saxena, Lala Li, Jay Whang, Emily~L
  Denton, Kamyar Ghasemipour, Raphael Gontijo~Lopes, Burcu Karagol~Ayan, Tim
  Salimans, et~al.
\newblock Photorealistic text-to-image diffusion models with deep language
  understanding.
\newblock {\em Advances in Neural Information Processing Systems},
  35:36479--36494, 2022.

\bibitem{xu2022groupvit}
Jiarui Xu, Shalini De~Mello, Sifei Liu, Wonmin Byeon, Thomas Breuel, Jan Kautz,
  and Xiaolong Wang.
\newblock Groupvit: Semantic segmentation emerges from text supervision.
\newblock In {\em Proceedings of the IEEE/CVF Conference on Computer Vision and
  Pattern Recognition}, pages 18134--18144, 2022.

\bibitem{gu2021open}
Xiuye Gu, Tsung-Yi Lin, Weicheng Kuo, and Yin Cui.
\newblock Open-vocabulary object detection via vision and language knowledge
  distillation.
\newblock {\em arXiv preprint arXiv:2104.13921}, 2021.

\bibitem{li2022grounded}
Liunian~Harold Li, Pengchuan Zhang, Haotian Zhang, Jianwei Yang, Chunyuan Li,
  Yiwu Zhong, Lijuan Wang, Lu~Yuan, Lei Zhang, Jenq-Neng Hwang, et~al.
\newblock Grounded language-image pre-training.
\newblock In {\em Proceedings of the IEEE/CVF Conference on Computer Vision and
  Pattern Recognition}, pages 10965--10975, 2022.

\bibitem{vinker2022clipasso}
Yael Vinker, Ehsan Pajouheshgar, Jessica~Y Bo, Roman~Christian Bachmann,
  Amit~Haim Bermano, Daniel Cohen-Or, Amir Zamir, and Ariel Shamir.
\newblock Clipasso: Semantically-aware object sketching.
\newblock {\em ACM Transactions on Graphics (TOG)}, 41(4):1--11, 2022.

\bibitem{luo2022clip4clip}
Huaishao Luo, Lei Ji, Ming Zhong, Yang Chen, Wen Lei, Nan Duan, and Tianrui Li.
\newblock Clip4clip: An empirical study of clip for end to end video clip
  retrieval and captioning.
\newblock {\em Neurocomputing}, 508:293--304, 2022.

\bibitem{wang2021actionclip}
Mengmeng Wang, Jiazheng Xing, and Yong Liu.
\newblock Actionclip: A new paradigm for video action recognition.
\newblock {\em arXiv preprint arXiv:2109.08472}, 2021.

\bibitem{jia2022visual}
Menglin Jia, Luming Tang, Bor-Chun Chen, Claire Cardie, Serge Belongie, Bharath
  Hariharan, and Ser-Nam Lim.
\newblock Visual prompt tuning.
\newblock In {\em Computer Vision--ECCV 2022: 17th European Conference, Tel
  Aviv, Israel, October 23--27, 2022, Proceedings, Part XXXIII}, pages
  709--727. Springer, 2022.

\bibitem{sohn2023visual}
Kihyuk Sohn, Huiwen Chang, Jos{\'e} Lezama, Luisa Polania, Han Zhang, Yuan Hao,
  Irfan Essa, and Lu~Jiang.
\newblock Visual prompt tuning for generative transfer learning.
\newblock In {\em Proceedings of the IEEE/CVF Conference on Computer Vision and
  Pattern Recognition}, pages 19840--19851, 2023.

\bibitem{deng2023segment}
Ruining Deng, Can Cui, Quan Liu, Tianyuan Yao, Lucas~W Remedios, Shunxing Bao,
  Bennett~A Landman, Lee~E Wheless, Lori~A Coburn, Keith~T Wilson, et~al.
\newblock Segment anything model (sam) for digital pathology: Assess zero-shot
  segmentation on whole slide imaging.
\newblock {\em arXiv preprint arXiv:2304.04155}, 2023.

\bibitem{mazurowski2023segment}
Maciej~A Mazurowski, Haoyu Dong, Hanxue Gu, Jichen Yang, Nicholas Konz, and
  Yixin Zhang.
\newblock Segment anything model for medical image analysis: an experimental
  study.
\newblock {\em arXiv preprint arXiv:2304.10517}, 2023.

\bibitem{wu2023medical}
Junde Wu, Rao Fu, Huihui Fang, Yuanpei Liu, Zhaowei Wang, Yanwu Xu, Yueming
  Jin, and Tal Arbel.
\newblock Medical sam adapter: Adapting segment anything model for medical
  image segmentation.
\newblock {\em arXiv preprint arXiv:2304.12620}, 2023.

\bibitem{zhou2023can}
Tao Zhou, Yizhe Zhang, Yi~Zhou, Ye~Wu, and Chen Gong.
\newblock Can sam segment polyps?
\newblock {\em arXiv preprint arXiv:2304.07583}, 2023.

\bibitem{shi2023generalist}
Peilun Shi, Jianing Qiu, Sai Mu~Dalike Abaxi, Hao Wei, Frank P-W Lo, and
  Wu~Yuan.
\newblock Generalist vision foundation models for medical imaging: A case study
  of segment anything model on zero-shot medical segmentation.
\newblock {\em Diagnostics}, 13(11):1947, 2023.

\bibitem{he2023accuracy}
Sheng He, Rina Bao, Jingpeng Li, P~Ellen Grant, and Yangming Ou.
\newblock Accuracy of segment-anything model (sam) in medical image
  segmentation tasks.
\newblock {\em arXiv preprint arXiv:2304.09324}, 2023.

\bibitem{zhang2023input}
Yizhe Zhang, Tao Zhou, Peixian Liang, and Danny~Z Chen.
\newblock Input augmentation with sam: Boosting medical image segmentation with
  segmentation foundation model.
\newblock {\em arXiv preprint arXiv:2304.11332}, 2023.

\bibitem{cheng2023segment}
Yangming Cheng, Liulei Li, Yuanyou Xu, Xiaodi Li, Zongxin Yang, Wenguan Wang,
  and Yi~Yang.
\newblock Segment and track anything.
\newblock {\em arXiv preprint arXiv:2305.06558}, 2023.

\bibitem{yang2023track}
Jinyu Yang, Mingqi Gao, Zhe Li, Shang Gao, Fangjing Wang, and Feng Zheng.
\newblock Track anything: Segment anything meets videos.
\newblock {\em arXiv preprint arXiv:2304.11968}, 2023.

\bibitem{yuan2023automated}
Andrew Yuan, Maura Sabatos-DeVito, Alexandra~L Bey, Samantha Major, Kimberly~LH
  Carpenter, Lauren Franz, Jill Howard, Saritha Vermeer, Ryan Simmons, Jesse
  Troy, et~al.
\newblock Automated movement tracking of young autistic children during free
  play is correlated with clinical features associated with autism.
\newblock {\em Autism}, page 13623613231169546, 2023.

\bibitem{cao2023ntire}
Mingdeng Cao, Chong Mou, Fanghua Yu, Xintao Wang, Yinqiang Zheng, Jian Zhang,
  Chao Dong, Gen Li, Ying Shan, Radu Timofte, et~al.
\newblock Ntire 2023 challenge on 360deg omnidirectional image and video
  super-resolution: Datasets, methods and results.
\newblock In {\em Proceedings of the IEEE/CVF Conference on Computer Vision and
  Pattern Recognition}, pages 1731--1745, 2023.

\bibitem{julka2023knowledge}
Sahib Julka and Michael Granitzer.
\newblock Knowledge distillation with segment anything (sam) model for
  planetary geological mapping.
\newblock {\em arXiv preprint arXiv:2305.07586}, 2023.

\bibitem{he2023scalable}
Haibin He, Jing Zhang, Mengyang Xu, Juhua Liu, Bo~Du, and Dacheng Tao.
\newblock Scalable mask annotation for video text spotting.
\newblock {\em arXiv preprint arXiv:2305.01443}, 2023.

\bibitem{he2023weakly}
Chunming He, Kai Li, Yachao Zhang, Guoxia Xu, Longxiang Tang, Yulun Zhang,
  Zhenhua Guo, and Xiu Li.
\newblock Weakly-supervised concealed object segmentation with sam-based pseudo
  labeling and multi-scale feature grouping.
\newblock {\em arXiv preprint arXiv:2305.11003}, 2023.

\bibitem{shen2023anything}
Qiuhong Shen, Xingyi Yang, and Xinchao Wang.
\newblock Anything-3d: Towards single-view anything reconstruction in the wild.
\newblock {\em arXiv preprint arXiv:2304.10261}, 2023.

\bibitem{wang2023sam}
An~Wang, Mobarakol Islam, Mengya Xu, Yang Zhang, and Hongliang Ren.
\newblock Sam meets robotic surgery: An empirical study in robustness
  perspective.
\newblock {\em arXiv preprint arXiv:2304.14674}, 2023.

\bibitem{diaz2023robot}
Vicente~Garc{\'\i}a D{\'\i}az, R~Dinesh~Jackson Samuel, Adhiyaman Manickam,
  Vijayalakshmi Saravanan, Ashish~Kr Luhach, and Sujatha Krishnamoorthy.
\newblock Robot based transurethral bladder tumor resection with automatic
  detection of tumor cells.
\newblock {\em Measurement}, 206:112079, 2023.

\bibitem{beauchat2023analyzing}
Tessa Beauchat, Yuqing Hu, Robert~M Leicht, and Clinton Suanico.
\newblock Analyzing schedule dependency and sequencing changes for robotic
  construction using graph analysis.
\newblock {\em Journal of Computing in Civil Engineering}, 37(1):04022043,
  2023.

\bibitem{yu2023inpaint}
Tao Yu, Runseng Feng, Ruoyu Feng, Jinming Liu, Xin Jin, Wenjun Zeng, and Zhibo
  Chen.
\newblock Inpaint anything: Segment anything meets image inpainting.
\newblock {\em arXiv preprint arXiv:2304.06790}, 2023.

\bibitem{roy2023sam}
Saikat Roy, Tassilo Wald, Gregor Koehler, Maximilian~R Rokuss, Nico Disch,
  Julius Holzschuh, David Zimmerer, and Klaus~H Maier-Hein.
\newblock Sam. md: Zero-shot medical image segmentation capabilities of the
  segment anything model.
\newblock {\em arXiv preprint arXiv:2304.05396}, 2023.

\bibitem{zhou2022learning}
Kaiyang Zhou, Jingkang Yang, Chen~Change Loy, and Ziwei Liu.
\newblock Learning to prompt for vision-language models.
\newblock {\em International Journal of Computer Vision}, 130(9):2337--2348,
  2022.

\bibitem{rao2022denseclip}
Yongming Rao, Wenliang Zhao, Guangyi Chen, Yansong Tang, Zheng Zhu, Guan Huang,
  Jie Zhou, and Jiwen Lu.
\newblock Denseclip: Language-guided dense prediction with context-aware
  prompting.
\newblock In {\em Proceedings of the IEEE/CVF Conference on Computer Vision and
  Pattern Recognition}, pages 18082--18091, 2022.

\bibitem{khattak2023maple}
Muhammad~Uzair Khattak, Hanoona Rasheed, Muhammad Maaz, Salman Khan, and
  Fahad~Shahbaz Khan.
\newblock Maple: Multi-modal prompt learning.
\newblock In {\em Proceedings of the IEEE/CVF Conference on Computer Vision and
  Pattern Recognition}, pages 19113--19122, 2023.

\bibitem{kawar2023imagic}
Bahjat Kawar, Shiran Zada, Oran Lang, Omer Tov, Huiwen Chang, Tali Dekel, Inbar
  Mosseri, and Michal Irani.
\newblock Imagic: Text-based real image editing with diffusion models.
\newblock In {\em Proceedings of the IEEE/CVF Conference on Computer Vision and
  Pattern Recognition}, pages 6007--6017, 2023.

\bibitem{tao2023galip}
Ming Tao, Bing-Kun Bao, Hao Tang, and Changsheng Xu.
\newblock Galip: Generative adversarial clips for text-to-image synthesis.
\newblock In {\em Proceedings of the IEEE/CVF Conference on Computer Vision and
  Pattern Recognition}, pages 14214--14223, 2023.

\bibitem{wang2023position}
Jinpeng Wang, Pan Zhou, Mike~Zheng Shou, and Shuicheng Yan.
\newblock Position-guided text prompt for vision-language pre-training.
\newblock In {\em Proceedings of the IEEE/CVF Conference on Computer Vision and
  Pattern Recognition}, pages 23242--23251, 2023.

\bibitem{xu2016deep}
Ning Xu, Brian Price, Scott Cohen, Jimei Yang, and Thomas~S Huang.
\newblock Deep interactive object selection.
\newblock In {\em Proceedings of the IEEE conference on computer vision and
  pattern recognition}, pages 373--381, 2016.

\bibitem{wang2018interactive}
Guotai Wang, Wenqi Li, Maria~A Zuluaga, Rosalind Pratt, Premal~A Patel, Michael
  Aertsen, Tom Doel, Anna~L David, Jan Deprest, S{\'e}bastien Ourselin, et~al.
\newblock Interactive medical image segmentation using deep learning with
  image-specific fine tuning.
\newblock {\em IEEE transactions on medical imaging}, 37(7):1562--1573, 2018.

\bibitem{jang2019interactive}
Won-Dong Jang and Chang-Su Kim.
\newblock Interactive image segmentation via backpropagating refinement scheme.
\newblock In {\em Proceedings of the IEEE/CVF Conference on Computer Vision and
  Pattern Recognition}, pages 5297--5306, 2019.

\bibitem{lin2020interactive}
Zheng Lin, Zhao Zhang, Lin-Zhuo Chen, Ming-Ming Cheng, and Shao-Ping Lu.
\newblock Interactive image segmentation with first click attention.
\newblock In {\em Proceedings of the IEEE/CVF conference on computer vision and
  pattern recognition}, pages 13339--13348, 2020.

\bibitem{chen2021conditional}
Xi~Chen, Zhiyan Zhao, Feiwu Yu, Yilei Zhang, and Manni Duan.
\newblock Conditional diffusion for interactive segmentation.
\newblock In {\em Proceedings of the IEEE/CVF International Conference on
  Computer Vision}, pages 7345--7354, 2021.

\bibitem{lempitsky2009image}
Victor Lempitsky, Pushmeet Kohli, Carsten Rother, and Toby Sharp.
\newblock Image segmentation with a bounding box prior.
\newblock In {\em 2009 IEEE 12th international conference on computer vision},
  pages 277--284. IEEE, 2009.

\bibitem{wu2014milcut}
Jiajun Wu, Yibiao Zhao, Jun-Yan Zhu, Siwei Luo, and Zhuowen Tu.
\newblock Milcut: A sweeping line multiple instance learning paradigm for
  interactive image segmentation.
\newblock In {\em Proceedings of the IEEE conference on computer vision and
  pattern recognition}, pages 256--263, 2014.

\bibitem{rajchl2016deepcut}
Martin Rajchl, Matthew~CH Lee, Ozan Oktay, Konstantinos Kamnitsas, Jonathan
  Passerat-Palmbach, Wenjia Bai, Mellisa Damodaram, Mary~A Rutherford, Joseph~V
  Hajnal, Bernhard Kainz, et~al.
\newblock Deepcut: Object segmentation from bounding box annotations using
  convolutional neural networks.
\newblock {\em IEEE transactions on medical imaging}, 36(2):674--683, 2016.

\bibitem{batra2010icoseg}
Dhruv Batra, Adarsh Kowdle, Devi Parikh, Jiebo Luo, and Tsuhan Chen.
\newblock icoseg: Interactive co-segmentation with intelligent scribble
  guidance.
\newblock In {\em 2010 IEEE computer society conference on computer vision and
  pattern recognition}, pages 3169--3176. IEEE, 2010.

\bibitem{bai2014error}
Junjie Bai and Xiaodong Wu.
\newblock Error-tolerant scribbles based interactive image segmentation.
\newblock In {\em Proceedings of the IEEE Conference on Computer Vision and
  Pattern Recognition}, pages 392--399, 2014.

\bibitem{lin2016scribblesup}
Di~Lin, Jifeng Dai, Jiaya Jia, Kaiming He, and Jian Sun.
\newblock Scribblesup: Scribble-supervised convolutional networks for semantic
  segmentation.
\newblock In {\em Proceedings of the IEEE conference on computer vision and
  pattern recognition}, pages 3159--3167, 2016.

\bibitem{shaban2017one}
Amirreza Shaban, Shray Bansal, Zhen Liu, Irfan Essa, and Byron Boots.
\newblock One-shot learning for semantic segmentation.
\newblock {\em arXiv preprint arXiv:1709.03410}, 2017.

\bibitem{dong2018few}
Nanqing Dong and Eric~P Xing.
\newblock Few-shot semantic segmentation with prototype learning.
\newblock In {\em BMVC}, volume~3, 2018.

\bibitem{wang2019panet}
Kaixin Wang, Jun~Hao Liew, Yingtian Zou, Daquan Zhou, and Jiashi Feng.
\newblock Panet: Few-shot image semantic segmentation with prototype alignment.
\newblock In {\em proceedings of the IEEE/CVF international conference on
  computer vision}, pages 9197--9206, 2019.

\bibitem{chen2022adaptformer}
Shoufa Chen, Chongjian Ge, Zhan Tong, Jiangliu Wang, Yibing Song, Jue Wang, and
  Ping Luo.
\newblock Adaptformer: Adapting vision transformers for scalable visual
  recognition.
\newblock {\em arXiv preprint arXiv:2205.13535}, 2022.

\bibitem{jie2022convolutional}
Shibo Jie and Zhi-Hong Deng.
\newblock Convolutional bypasses are better vision transformer adapters.
\newblock {\em arXiv preprint arXiv:2207.07039}, 2022.

\bibitem{zhu2023visual}
Jiawen Zhu, Simiao Lai, Xin Chen, Dong Wang, and Huchuan Lu.
\newblock Visual prompt multi-modal tracking.
\newblock In {\em Proceedings of the IEEE/CVF Conference on Computer Vision and
  Pattern Recognition}, pages 9516--9526, 2023.

\bibitem{huang2023diversity}
Qidong Huang, Xiaoyi Dong, Dongdong Chen, Weiming Zhang, Feifei Wang, Gang Hua,
  and Nenghai Yu.
\newblock Diversity-aware meta visual prompting.
\newblock In {\em Proceedings of the IEEE/CVF Conference on Computer Vision and
  Pattern Recognition}, pages 10878--10887, 2023.

\bibitem{jain2023oneformer}
Jitesh Jain, Jiachen Li, Mang~Tik Chiu, Ali Hassani, Nikita Orlov, and Humphrey
  Shi.
\newblock Oneformer: One transformer to rule universal image segmentation.
\newblock In {\em Proceedings of the IEEE/CVF Conference on Computer Vision and
  Pattern Recognition}, pages 2989--2998, 2023.

\bibitem{wang2023seggpt}
Xinlong Wang, Xiaosong Zhang, Yue Cao, Wen Wang, Chunhua Shen, and Tiejun
  Huang.
\newblock Seggpt: Segmenting everything in context.
\newblock {\em arXiv preprint arXiv:2304.03284}, 2023.

\bibitem{zou2023segment}
Xueyan Zou, Jianwei Yang, Hao Zhang, Feng Li, Linjie Li, Jianfeng Gao, and
  Yong~Jae Lee.
\newblock Segment everything everywhere all at once.
\newblock {\em arXiv preprint arXiv:2304.06718}, 2023.

\bibitem{li2023uni}
Hao Li, Jinguo Zhu, Xiaohu Jiang, Xizhou Zhu, Hongsheng Li, Chun Yuan, Xiaohua
  Wang, Yu~Qiao, Xiaogang Wang, Wenhai Wang, et~al.
\newblock Uni-perceiver v2: A generalist model for large-scale vision and
  vision-language tasks.
\newblock In {\em Proceedings of the IEEE/CVF Conference on Computer Vision and
  Pattern Recognition}, pages 2691--2700, 2023.

\bibitem{ji2023segment}
Wei Ji, Jingjing Li, Qi~Bi, Wenbo Li, and Li~Cheng.
\newblock Segment anything is not always perfect: An investigation of sam on
  different real-world applications.
\newblock {\em arXiv preprint arXiv:2304.05750}, 2023.

\bibitem{ma2023can}
Zhiheng Ma, Xiaopeng Hong, and Qinnan Shangguan.
\newblock Can sam count anything? an empirical study on sam counting.
\newblock {\em arXiv preprint arXiv:2304.10817}, 2023.

\bibitem{zhang2023text2seg}
Jielu Zhang, Zhongliang Zhou, Gengchen Mai, Lan Mu, Mengxuan Hu, and Sheng Li.
\newblock Text2seg: Remote sensing image semantic segmentation via text-guided
  visual foundation models.
\newblock {\em arXiv preprint arXiv:2304.10597}, 2023.

\bibitem{wang2023scaling}
Di~Wang, Jing Zhang, Bo~Du, Dacheng Tao, and Liangpei Zhang.
\newblock Scaling-up remote sensing segmentation dataset with segment anything
  model.
\newblock {\em arXiv preprint arXiv:2305.02034}, 2023.

\bibitem{chen2023sam}
Tianrun Chen, Lanyun Zhu, Chaotao Ding, Runlong Cao, Shangzhan Zhang, Yan Wang,
  Zejian Li, Lingyun Sun, Papa Mao, and Ying Zang.
\newblock Sam fails to segment anything?--sam-adapter: Adapting sam in
  underperformed scenes: Camouflage, shadow, and more.
\newblock {\em arXiv preprint arXiv:2304.09148}, 2023.

\bibitem{wang2023caption}
Teng Wang, Jinrui Zhang, Junjie Fei, Yixiao Ge, Hao Zheng, Yunlong Tang, Zhe
  Li, Mingqi Gao, Shanshan Zhao, Ying Shan, et~al.
\newblock Caption anything: Interactive image description with diverse
  multimodal controls.
\newblock {\em arXiv preprint arXiv:2305.02677}, 2023.

\bibitem{cao2023segment}
Yunkang Cao, Xiaohao Xu, Chen Sun, Yuqi Cheng, Zongwei Du, Liang Gao, and
  Weiming Shen.
\newblock Segment any anomaly without training via hybrid prompt
  regularization.
\newblock {\em arXiv preprint arXiv:2305.10724}, 2023.

\bibitem{xie2023edit}
Defeng Xie, Ruichen Wang, Jian Ma, Chen Chen, Haonan Lu, Dong Yang, Fobo Shi,
  and Xiaodong Lin.
\newblock Edit everything: A text-guided generative system for images editing.
\newblock {\em arXiv preprint arXiv:2304.14006}, 2023.

\bibitem{sun2023explain}
Ao~Sun, Pingchuan Ma, Yuanyuan Yuan, and Shuai Wang.
\newblock Explain any concept: Segment anything meets concept-based
  explanation.
\newblock {\em arXiv preprint arXiv:2305.10289}, 2023.

\bibitem{liu2022p}
Xiao Liu, Kaixuan Ji, Yicheng Fu, Weng Tam, Zhengxiao Du, Zhilin Yang, and Jie
  Tang.
\newblock P-tuning: Prompt tuning can be comparable to fine-tuning across
  scales and tasks.
\newblock In {\em Proceedings of the 60th Annual Meeting of the Association for
  Computational Linguistics (Volume 2: Short Papers)}, pages 61--68, 2022.

\bibitem{hu2022p3}
Xiaomeng Hu, Shi Yu, Chenyan Xiong, Zhenghao Liu, Zhiyuan Liu, and Ge~Yu.
\newblock P3 ranker: Mitigating the gaps between pre-training and ranking
  fine-tuning with prompt-based learning and pre-finetuning.
\newblock In {\em Proceedings of the 45th International ACM SIGIR Conference on
  Research and Development in Information Retrieval}, pages 1956--1962, 2022.

\bibitem{abdel2022dialect}
Reem Abdel-Salam.
\newblock Dialect \& sentiment identification in nuanced arabic tweets using an
  ensemble of prompt-based, fine-tuned, and multitask bert-based models.
\newblock In {\em Proceedings of the The Seventh Arabic Natural Language
  Processing Workshop (WANLP)}, pages 452--457, 2022.

\bibitem{matsuo2022deep}
Yutaka Matsuo, Yann LeCun, Maneesh Sahani, Doina Precup, David Silver, Masashi
  Sugiyama, Eiji Uchibe, and Jun Morimoto.
\newblock Deep learning, reinforcement learning, and world models.
\newblock {\em Neural Networks}, 152:267--275, 2022.

\bibitem{ladosz2022exploration}
Pawel Ladosz, Lilian Weng, Minwoo Kim, and Hyondong Oh.
\newblock Exploration in deep reinforcement learning: A survey.
\newblock {\em Information Fusion}, 85:1--22, 2022.

\bibitem{wurman2022outracing}
Peter~R Wurman, Samuel Barrett, Kenta Kawamoto, James MacGlashan, Kaushik
  Subramanian, Thomas~J Walsh, Roberto Capobianco, Alisa Devlic, Franziska
  Eckert, Florian Fuchs, et~al.
\newblock Outracing champion gran turismo drivers with deep reinforcement
  learning.
\newblock {\em Nature}, 602(7896):223--228, 2022.

\bibitem{morse2022determinants}
WH~Morse and RT~Kelleher.
\newblock Determinants of reinforcement and punishment.
\newblock In {\em Handbook of operant behavior}, pages 174--200. Routledge,
  2022.

\bibitem{thomas2022efficient}
Bethan Thomas, Samuel Kessler, and Salah Karout.
\newblock Efficient adapter transfer of self-supervised speech models for
  automatic speech recognition.
\newblock In {\em ICASSP 2022-2022 IEEE International Conference on Acoustics,
  Speech and Signal Processing (ICASSP)}, pages 7102--7106. IEEE, 2022.

\bibitem{goel2022cross}
Divyam Goel, Ramansh Grover, and Fatemeh~H Fard.
\newblock On the cross-modal transfer from natural language to code through
  adapter modules.
\newblock In {\em Proceedings of the 30th IEEE/ACM International Conference on
  Program Comprehension}, pages 71--81, 2022.

\bibitem{zhao2022decoupled}
Borui Zhao, Quan Cui, Renjie Song, Yiyu Qiu, and Jiajun Liang.
\newblock Decoupled knowledge distillation.
\newblock In {\em Proceedings of the IEEE/CVF Conference on computer vision and
  pattern recognition}, pages 11953--11962, 2022.

\bibitem{lin2022knowledge}
Sihao Lin, Hongwei Xie, Bing Wang, Kaicheng Yu, Xiaojun Chang, Xiaodan Liang,
  and Gang Wang.
\newblock Knowledge distillation via the target-aware transformer.
\newblock In {\em Proceedings of the IEEE/CVF Conference on Computer Vision and
  Pattern Recognition}, pages 10915--10924, 2022.

\bibitem{chen2022knowledge}
Defang Chen, Jian-Ping Mei, Hailin Zhang, Can Wang, Yan Feng, and Chun Chen.
\newblock Knowledge distillation with the reused teacher classifier.
\newblock In {\em Proceedings of the IEEE/CVF conference on computer vision and
  pattern recognition}, pages 11933--11942, 2022.

\bibitem{xie2023towards}
Lingxi Xie, Longhui Wei, Xiaopeng Zhang, Kaifeng Bi, Xiaotao Gu, Jianlong
  Chang, and Qi~Tian.
\newblock Towards agi in computer vision: Lessons learned from gpt and large
  language models.
\newblock {\em arXiv preprint arXiv:2306.08641}, 2023.

\bibitem{moravec1988mind}
Hans Moravec.
\newblock {\em Mind children: The future of robot and human intelligence}.
\newblock Harvard University Press, 1988.

\bibitem{brooks1990elephants}
Rodney~A Brooks.
\newblock Elephants don't play chess.
\newblock {\em Robotics and autonomous systems}, 6(1-2):3--15, 1990.

\bibitem{ma2023segment}
Jun Ma and Bo~Wang.
\newblock Segment anything in medical images.
\newblock {\em arXiv preprint arXiv:2304.12306}, 2023.

\bibitem{lu2023agi}
Guoyu Lu, Sheng Li, Gengchen Mai, Jin Sun, Dajiang Zhu, Lilong Chai, Haijian
  Sun, Xianqiao Wang, Haixing Dai, Ninghao Liu, et~al.
\newblock Agi for agriculture.
\newblock {\em arXiv preprint arXiv:2304.06136}, 2023.

\bibitem{yang2023sam}
Xiao Yang, Haixing Dai, Zihao Wu, Ramesh Bist, Sachin Subedi, Jin Sun, Guoyu
  Lu, Changying Li, Tianming Liu, and Lilong Chai.
\newblock Sam for poultry science.
\newblock {\em arXiv preprint arXiv:2305.10254}, 2023.

\end{thebibliography}

\end{CJK}
\end{document}